\documentclass[smallextended,natbib]{svjour3} 

\newif\ifdraft

\usepackage{blindtext}

\usepackage{amssymb}
\usepackage{amsmath}
\usepackage[lined,boxed,linesnumbered]{algorithm2e}
\usepackage{array}
\usepackage{color}
\usepackage{colortbl}
\usepackage{dcolumn}
\usepackage{graphicx}
\usepackage{hyperref}
\usepackage{latexsym} 
\usepackage{multicol}
\usepackage{multirow}
\usepackage{rotating}
\usepackage{soul}
\usepackage{url}
\usepackage[usenames]{xcolor}
\usepackage[utf8]{inputenc}
\usepackage{tabularx,booktabs}

\usepackage{etoolbox}

\newcommand{\fabscomment}[1]{\ifdraft{\leavevmode\color{cyan}[FS]: {#1}}\else{\vspace{0ex}}\fi}

\newcommand{\blue}[1]{\ifdraft{\leavevmode\color{blue}{#1}}\else{\leavevmode\color{black}{#1}}\fi}
\newcommand{\no}{\cellcolor[gray]{.8}No}
\newcommand{\gr}{\cellcolor[gray]{.8}}
\newcommand{\side}[1]{\begin{sideways}{#1}\end{sideways}}

\DeclareMathOperator{\aae}{AE}
\DeclareMathOperator{\nae}{NAE}
\DeclareMathOperator{\rae}{RAE}
\DeclareMathOperator{\nrae}{NRAE}
\DeclareMathOperator{\kld}{KLD}
\DeclareMathOperator{\nkld}{NKLD}

\DeclareMathOperator{\pd}{PD}

\DeclareMathOperator{\se}{SE}
\DeclareMathOperator{\nse}{NSE}
\DeclareMathOperator{\cra}{CR}

\DeclareMathOperator{\dr}{DR}

\DeclareMathOperator{\nas}{NAS}
\DeclareMathOperator{\nss}{NSS}

\newtheorem{ex}{Example}

\newcounter{propcount}
\newtheorem{prop}{\refstepcounter{propcount}Property}

\newcounter{propprovcount}
\newcolumntype{Y}{>{\centering\arraybackslash}X}

\journalname{Information Retrieval Journal}

\begin{document}

\title{Evaluation Measures for Quantification: An Axiomatic Approach} \titlerunning{Evaluation Measures for
Quantification} \author{Fabrizio Sebastiani} \authorrunning{F.\
Sebastiani} \institute{Fabrizio
Sebastiani \at Istituto di Scienza e Tecnologie dell'Informazione \\
Consiglio Nazionale delle Ricerche \\ 56124 Pisa, Italy \\
\email{fabrizio.sebastiani@isti.cnr.it}} \date{Received: \today /
Accepted: DD MM YYYY}

\maketitle


\begin{abstract}
  Quantification is the task of estimating, given a set $\sigma$ of
  unlabelled items and a set of classes
  $\mathcal{C}=\{c_{1}, \ldots, c_{|\mathcal{C}|}\}$, the prevalence
  (or ``relative frequency'') in $\sigma$ of each class
  $c_{i}\in \mathcal{C}$. While quantification may in principle be
  solved by classifying each item in $\sigma$ and counting how many
  such items have been labelled with $c_{i}$, it has long been shown
  that this ``classify and count'' (CC) method yields suboptimal
  quantification accuracy. As a result, quantification is no longer
  considered a mere byproduct of classification, and has evolved as a
  task of its own. While the scientific community has devoted a lot of
  attention to devising more accurate quantification methods, it has
  not devoted much to discussing what properties an \emph{evaluation
  measure for quantification} (EMQ) should enjoy, and which EMQs
  should be adopted as a result.  This paper lies down a number of
  interesting properties that an EMQ may or may not enjoy, discusses
  if (and when) each of these properties is desirable, surveys the
  EMQs that have been used so far, and discusses whether they enjoy or
  not the above properties. As a result of this investigation, some of
  the EMQs that have been used in the literature turn out to be
  severely unfit, while others emerge as closer to what the
  quantification community actually needs. However, a significant
  result is that no existing EMQ satisfies all the properties
  identified as desirable, thus indicating that more research is
  needed in order to identify (or synthesize) a truly adequate
  EMQ. \fabscomment{Checked, Sep 1.}
\end{abstract}


%
%


\section{Introduction}
\label{sec:introduction}

\noindent \emph{Quantification} (also known as ``supervised prevalence
estimation'' \citep{Barranquero:2013fk}, or ``class prior estimation''
\citep{Plessis:2017sp}) is the task of estimating, given a set
$\sigma$ of unlabelled items and a set of classes
$\mathcal{C}=\{c_{1}, \ldots, c_{|\mathcal{C}|}\}$, the relative
frequency (or ``prevalence'') $p(c_{i})$ of each class
$c_{i}\in \mathcal{C}$, i.e., the fraction of items in $\sigma$ that
belong to $c_{i}$. When each item belongs to exactly one class, since
$0\leq p(c_{i})\leq 1$ and $\sum_{c_{i}\in\mathcal{C}}p(c_{i})=1$, $p$
is a \emph{distribution} of the items in $\sigma$ across the classes
in $\mathcal{C}$ (the \emph{true distribution}), and quantification
thus amounts to estimating $p$ (i.e., to computing a \emph{predicted
distribution} $\hat{p}$).

Quantification is important in many disciplines (such as e.g., market
research, political science, the social sciences, and epidemiology)
which usually deal with aggregate (as opposed to individual) data. In
these contexts, classifying individual unlabelled instances is usually
not a primary goal, while estimating the prevalence of the classes of
interest in the data is. For instance, when classifying the tweets
about a certain entity (e.g., a political candidate) as displaying
either a \textsf{Positive} or a \textsf{Negative} stance towards the
entity, we are usually not much interested in the class of a specific
tweet: instead, we usually want to know the fraction of these tweets
that belong to the class \citep{Gao:2016uq}.

Quantification may in principle be solved via classification, i.e., by
classifying each item in $\sigma$ and counting, for all
$c_{i}\in \mathcal{C}$, how many such items have been labelled with
$c_{i}$. However, it has been shown in a multitude of works (see e.g.,
\citep{Barranquero:2015fr,Bella:2010kx,Esuli:2015gh,Forman:2008kx,Gao:2016uq,Hopkins:2010fk})
that this ``classify and count'' (CC) method yields suboptimal
quantification accuracy. Simply put, the reason of this suboptimality
is that most classifiers are optimized for classification accuracy,
and not for quantification accuracy. These two notions do not
coincide, since the former is, by and large, inversely proportional to
the sum $(FP_{i}+FN_{i})$ of the false positives and the false
negatives for $c_{i}$ in the contingency table, while the latter is,
by and large, inversely proportional to the absolute difference
$|FP_{i}-FN_{i}|$ of the two. As a result, quantification has come to
be no longer considered a mere byproduct of classification, and has
evolved as a task of its own, devoted to designing methods and
algorithms that deliver better prevalence estimates than CC (see
\citep{Gonzalez:2017it} for a survey of methods and results).

While the scientific community working on quantification has devoted a
lot of attention to devising new and more accurate quantification
methods, it has not devoted much to discussing how quantification
accuracy should be measured, i.e., what properties an \emph{evaluation
measure for quantification} (EMQ) should enjoy, and which EMQs should
be adopted as a result. Sometimes, new EMQs have been introduced
without arguing why they are supposedly better than existing ones. As
a result, there is no consensus (and, what is worse: no debate) in the
field as to which EMQ (if any) is the best. Different authors use
different EMQs without properly justifying their choice, and the
consequence is that different results, even when obtained on the same
dataset, are not comparable. Even worse, it may be the case that an
improvement, sanctioned by an ``inappropriate'' EMQ, obtained by a
newly proposed method with respect to a baseline, may correspond to no
real improvement when measured according to an ``appropriate'' EMQ.

This paper attempts to shed some light on the issue of which
evaluation measure(s) should be used for quantification. In order to
do so, we (a) lie down a number of interesting properties that an EMQ
may or may not enjoy, (b) discuss whether (or when) each of these
properties is desirable, (c) survey the EMQs that have been used so
far, and (d) discuss whether they enjoy or not the above
properties. As a result of this investigation, some of the EMQs that
have been used in the literature turn out to be severely unfit, while
others emerge as closer to ``what the quantification community
actually needs''. However, a significant result is that no existing
measure satisfies all the properties identified as desirable, thus
indicating that more research is needed in order to identify (or
synthesize) a truly adequate EMQ.

This paper follows in the tradition of the so-called ``axiomatic''
approach to ``evaluating evaluation'' in information retrieval (see
e.g.,
\citep{Amigo:2011fk,Busin:2013dw,Ferrante:2015qy,Ferrante:2018fk,Moffat:2013kx,Sebastiani:2015zl}),
which is based on describing (and often: arguing in favour of) a
number of properties (that most of this literature calls -- perhaps
improperly -- ``axioms'') that an evaluation measure for the task
being considered should intuitively satisfy. The benefit of this
approach is that it shifts the discussion from the evaluation measures
to their properties, which amounts to shifting the discussion from a
complex construction to its building blocks: once the scientific
community has agreed on a set of properties (the building blocks), it
then follows whether a given measure (the construction) is
satisfactory or not.

The paper is structured as follows. In Section
\ref{sec:singlelabelevaluationmeasures} we set the stage and define
the scope of our investigation. In Section \ref{sec:propertiesforSLQ}
we formally discuss properties that may or may not characterize an
EMQ, and argue if and when it is desirable that an EMQ enjoys them. In
Section \ref{sec:measures} we turn to examining the actual measures
that have been proposed or used in the quantification literature, and
discuss whether they comply or not with the properties introduced in
Section \ref{sec:propertiesforSLQ}. Section \ref{sec:discussion}
critically reexamines the results of Section \ref{sec:measures}, while
Section \ref{sec:conclusion} concludes, discussing aspects that the
present work still leaves open and avenues for further research.
\fabscomment{Checked, July 4.}


\section{Evaluating Single-Label Quantification}
\label{sec:singlelabelevaluationmeasures}

\noindent Let us fix some notation. Symbols $\sigma$, $\sigma'$,
$\sigma''$, \ldots will each denote a \emph{sample}, i.e., a nonempty
set of unlabelled items, while symbols $\mathcal{C}$, $\mathcal{C}'$,
$\mathcal{C}''$, \ldots will each denote a nonempty set of classes (or
\emph{codeframe}) across which the unlabelled items in a sample are
distributed. Symbols $c$, $c_{1}$, $c_{2}$, \ldots will each denote an
individual class. Given a class $c_{i}$, we will denote by
$\sigma_{i}$ the set of items in $\sigma$ that belong to $c_{i}$; we
will also denote by $|\sigma|$, $|\sigma'|$, $|\sigma''|$, \ldots the
number of items contained in samples $\sigma$, $\sigma'$, $\sigma''$,
\ldots . Symbols $p$, $p'$, $p''$ \ldots, will each denote a
\emph{true distribution} of the unlabelled items in a sample $\sigma$
across a codeframe $\mathcal{C}$, while symbols $\hat{p}$, $\hat{p}'$,
$\hat{p}''$, \ldots will each denote a \emph{predicted distribution}
(or \emph{estimator}), i.e., the result of estimating a true
distribution;\footnote{Consistently with most mathematical literature,
we use the caret symbol (\^\/\/) to indicate estimation.} symbol
$\mathcal{P}$ will denote the (infinite) set of all distributions on
$\mathcal{C}$.\footnote{In order to keep things simple we avoid
overspecifying the notation, thus leaving some aspects of it implicit;
e.g., in order to indicate a true distribution $p$ of the unlabelled
items in a sample $\sigma$ across a codeframe $\mathcal{C}$ we will
simply write $p$ instead of the more cumbersome
$p_{\sigma}^{\mathcal{C}}$, thus letting $\sigma$ and $\mathcal{C}$ be
inferred from context.} Finally, symbols $D$, $D'$, $D''$, \ldots will
each denote an EMQ, while symbols $\pi$, $\pi'$, $\pi''$, \ldots will
denote properties that an EMQ may enjoy or not.

Similarly to classification, there are different quantification
problems of applicative interest, based (a) on how many classes
codeframe $\mathcal{C}$ contains, and (b) how many of the classes in
$\mathcal{C}$ can be legitimately attributed to the same item. We
characterize quantification problems as follows:
\begin{enumerate}

\item \emph{Single-label quantification} (SLQ) is defined as
  quantification when each item belongs to exactly one of the classes
  in $\mathcal{C}=\{c_{1}, ..., c_{|\mathcal{C}|}\}$.
  
\item \emph{Multi-label quantification} (MLQ) is defined as
  quantification when the same item may belong to any number of
  classes (zero, one, or several) in
  $\mathcal{C}=\{c_{1}, ..., c_{|\mathcal{C}|}\}$.
   
\item \emph{Binary quantification} (BQ) may alternatively be defined

  \begin{enumerate}

  \item \label{item:BinAsSLQ} as SLQ with $|\mathcal{C}|=2$ (in this
    case $\mathcal{C}=\{c_{1},c_{2}\}$ and each item must belong to
    either $c_{1}$ or $c_{2}$), or

  \item \label{item:BinAsMLQ} as MLQ with $|\mathcal{C}|=1$ (in this
    case $\mathcal{C}=\{c\}$ and each item either belongs or does not
    belong to $c$).

  \end{enumerate}

\end{enumerate}

\noindent Since BQ is a special case of SLQ (see bullet
\ref{item:BinAsSLQ} above), any evaluation measure for SLQ is also an
evaluation measure for BQ. Likewise, any evaluation measure for BQ is
also an evaluation measure for MLQ, since evaluating a multi-label
\emph{quantifier} (i.e., a software artifact that estimates class
prevalences) is trivially equivalent to evaluating $|\mathcal{C}|$
binary quantifiers, one for each $c_{i}\in \mathcal{C}$. As a
consequence, in this paper we focus on the evaluation of SLQ, knowing
that all the solutions we discuss for SLQ also apply to BQ and
MLQ.\footnote{\blue{In this paper we do not discuss the evaluation of
\emph{ordinal quantification} (OQ), defined as SLQ with a codeframe
$\mathcal{C}=\{c_{1}, ..., c_{|\mathcal{C}|}\}$ on which a total order
$c_{1} \prec ... \prec c_{|\mathcal{C}|}$ is defined. Aside from
reasons of space, the reasons for disregarding OQ is that there has
been very little work on it (the only papers we know being
\citep{DaSanMartino:2016jk,DaSanMartino:2016ty,SemEval:2016:task4:ISTI-CNR}),
and that only one measure for OQ (the \emph{Earth Mover's Distance} --
see \citep{Esuli:2010fk}) has been proposed and used so far. For the
same reasons we do not discuss \emph{regression quantification} (RQ),
the task that stands to metric regression as single-label
quantification stands to single-label classification. RQ has been
studied even less than OQ, the only work appeared on this theme so far
being, to the best of our knowledge, \citep{Bella:2014kp}, which as an
evaluation measure has proposed the \emph{Cramér-von-Mises
$u$-statistic} (see \citep{Bella:2014kp} for details).}}


As already discussed, given a sample $\sigma$ of items
(single-)labelled according to
$\mathcal{C}=\{c_{1}, ..., c_{|\mathcal{C}|}\}$, quantification has to
do with determining, for each $c_{i}\in \mathcal{C}$, the fraction
$|\sigma_{i}|/|\sigma|$ of items in $\sigma$ that are labelled by
$c_{i}$. These $|\mathcal{C}|$ fractions actually form a distribution
$p$ of the items in $\sigma$ across the classes in $\mathcal{C}$;
quantification may thus be seen as generating a predicted distribution
$\hat{p}(c)$ over $\mathcal{C}$ that approximates a true distribution
$p(c)$ over $\mathcal{C}$. Evaluating quantification thus means
measuring how well $\hat{p}(c)$ fits $p(c)$. We will thus be concerned
with discussing the properties that a function that attempts to
measure this goodness-of-fit should enjoy; we hereafter use the
notation $D(p,\hat{p})$ to indicate such a function.\footnote{Note
that two distributions $p(c)$ and $\hat{p}(c)$ over $\mathcal{C}$ are
essentially two nonnegative-valued, length-normalized vectors of
dimensionality $|\mathcal{C}|$. The literature on EMQs thus obviously
intersects the literature on functions for computing the similarity of
two vectors.}

In this paper we assume that the EMQs we are concerned with are
measures of quantification error, and not of quantification
accuracy. The reason for this is that most, if not all, the EMQs that
have been used so far are indeed measures of error, so it would be
slightly unnatural to discuss our properties with reference to
quantification accuracy. Since any measure of accuracy can be turned
into a measure of error (typically: by taking its negation), this is
an inessential factor anyway.

\section{Properties for SLQ Error Measures}
\label{sec:propertiesforSLQ}


\subsection{Seven Desirable Properties}
\label{sec:the8properties}

\noindent In this section we examine a number of specific properties
that, as we argue, an EMQ should enjoy.
The spirit of our discussion will be essentially \emph{normative},
i.e., we will argue whether an EMQ should or should not enjoy a given
property, and whether this should hold regardless of the intended
application.  This is different, e.g., from the spirit of
\citep{Amigo:2011fk} (a work on the properties of evaluation measures
for document filtering), which has a \emph{descriptive} intent, i.e.,
describes a number of properties that such evaluation measures may or
may not enjoy but does not necessarily argue that all measures should
satisfy them.
 

The first four properties for EMQs that we discuss concern both
mathematical ``well-formedness'' and ease of interpretation.

\begin{prop}
  \label{ax:ioi}
  \textbf{Identity of Indiscernibles} (\textbf{IoI}).  For each
  codeframe $\mathcal{C}$, true distribution $p$, and predicted
  distribution $\hat{p}$, it holds that $D(p,\hat{p})=0$ if and only
  if $\hat{p}=p$.  \hfill \mbox{} \qed
\end{prop}

\begin{prop}
  \label{ax:nn}
  \textbf{Non-Negativity} (\textbf{NN}).  For each codeframe
  $\mathcal{C}$, true distribution $p$, and predicted distribution
  $\hat{p}$, it holds that $D(p,\hat{p})\geq0$.  \hfill \mbox{} \qed
\end{prop}

\noindent Imposing that an EMQ enjoys \textbf{IoI} and \textbf{NN} is
reasonable, since altogether they indicate a score for the
\emph{perfect estimator} (defined as the estimator $\hat{p}$ such that
$\hat{p}=p$) and stipulate that any other (non-perfect) estimator must
obtain a score strictly higher than it; both prescriptions fit our
understanding of $D$ as a measure or error. In mathematics, a function
of two probability distributions that enjoys \textbf{IoI} and
\textbf{NN} (two properties that, together, are often called
\emph{Positive Definiteness}) is called a \emph{divergence} (a.k.a.\
``contrast function'').\footnote{A divergence is often indicated by
the notation $D(p||\hat{p})$); we will prefer the more neutral
notation $D(p,\hat{p})$. Note also that a divergence can take as
arguments any two distributions $p$ and $q$ defined on the same space
of events, i.e., $p$ and $q$ need not be a true distribution and a
predicted distribution. However, since we will consider divergences
only as measures of fit between a true distribution and a predicted
distribution, we will use the more specific notation $D(p,\hat{p})$
rather than the more general $D(p,q)$.}

\begin{prop}
  \label{ax:mon}
  \textbf{Strict Monotonicity} (\textbf{MON}).  For each codeframe
  $\mathcal{C}$ and true distribution $p$, if there are predicted
  distributions $\hat{p}',\hat{p}''$ and classes
  $c_{1},c_{2}\in\mathcal{C}$ such that $\hat{p}'$ and $\hat{p}''$
  only differ for the fact that
  $\hat{p}''(c_{1})<\hat{p}'(c_{1})\leq p(c_{1})$ and
  $\hat{p}''(c_{2})>\hat{p}'(c_{2})\geq p(c_{2})$, with
  $|\hat{p}''(c_{1})-\hat{p}'(c_{1})| =
  |\hat{p}''(c_{2})-\hat{p}'(c_{2})|$, then it holds that
  $D(p,\hat{p}')<D(p,\hat{p}'')$. \hfill \mbox{} \qed
\end{prop}

\noindent If $D$ satisfies \textbf{MON}, this means that, all other
things being equal, a higher prediction error on a class $c_{1}$
(obviously matched by a higher prediction error, of opposite sign, on
another class $c_{2}$) implies a higher quantification error as
measured by $D$.

\begin{prop}
  \label{ax:max}
  \textbf{Maximum} (\textbf{MAX}).  There is a real value $\beta>0$
  such that, for each codeframe $\mathcal{C}$ and for each true
  distribution $p$, (i) there is a predicted distribution
  $\hat{p}^{*}$ such that $D(p,\hat{p}^{*})=\beta$, and (ii) for no
  predicted distribution $\hat{p}$ it holds that $D(p,\hat{p})>\beta$.
  \hfill \mbox{} \qed
\end{prop}

\noindent An estimator $\hat{p}^{*}$ that is the worst possible
estimator of $p$ for $D$ (i.e.,
$\hat{p}^{*}=\arg\max_{\hat{p}\in \mathcal{P}}D(p,\hat{p})$) will be
called the \emph{perverse estimator} of $p$ for $D$.  If $D$ satisfies
\textbf{MAX} and $\hat{p}^{*}$ is the perverse estimator of $p$ for
$D$, then $D(p,\hat{p}^{*})=\beta$.  Without loss of generality, in
the rest of this paper we will assume $\beta=1$; this assumption is
unproblematic since any interval $[0, \beta]$ can be rescaled to the
$[0,1]$ interval.

Altogether, these first four properties state (among other things)
that the range of an EMQ that satisfies them is \emph{independent of
the problem setting} (i.e., of $\mathcal{C}$, of its cardinality
$|\mathcal{C}|$, and of the true distribution $p$).\footnote{By the
``range'' of an EMQ here we actually mean its \emph{image} (i.e., the
set of values that the EMQ actually takes for its admissible input
values), and not just its codomain.} This is important, since in order
to be able to \emph{easily} judge whether a given value of $D$ means
high or low quantification error, not only we need to know what values
$D$ ranges on, but we need to know that these values are always the
same. In other words, should this range depend on $\mathcal{C}$, or on
its cardinality, or on the true distribution $p$, we would not be able
to easily interpret the meaning of a given value of $D$.

An additional, possibly even more important reason for requiring this
range to be independent of the problem setting is that, in order to
test a given quantification method, the EMQ usually needs to be
evaluated on a set of $n$ test samples $\sigma_{1}, ..., \sigma_{n}$
(each characterized by its own true distribution), and a measure of
central tendency (typically: the average or the median) across the $n$
resulting EMQ values then needs to be computed. If, for these $n$
samples, the EMQ ranges on $n$ different intervals, this measure of
central tendency will return unreliable results, since the results
obtained on the samples characterized by the wider such intervals will
exert a higher influence on the resulting value.

The fifth property we discuss deals with the relative impact of
underprediction and overprediction.

\begin{prop}
  \label{ax:imp}
  \textbf{Impartiality} (\textbf{IMP}).  For any codeframe
  $\mathcal{C}=\{c_{1},\ldots, c_{|\mathcal{C}|}\}$, true distribution
  $p$, predicted distributions $\hat{p}'$ and $\hat{p}''$, classes
  $c_{1},c_{2}\in \mathcal{C}$, and constant $a\geq 0$ such that
  $\hat{p}'$ and $\hat{p}''$ only differ for the fact that
  $\hat{p}'(c_{1})=p(c_{1})+a$, $\hat{p}'(c_{2})=p(c_{2})-a$,
  $\hat{p}''(c_{1})=p(c_{1})-a$, $\hat{p}''(c_{2})=p(c_{2})+a$, it
  holds that $D(p,\hat{p}')=D(p,\hat{p}'')$.  \hfill \mbox{} \qed
\end{prop}

\noindent In a nutshell, for an EMQ $D$ that enjoys \textbf{IMP},
underestimating a true prevalence $p(c)$ by an amount $a$ or
overestimating it by the same amount $a$ are equally serious
mistakes. For instance, assume that $\mathcal{C}=\{c_{1},c_{2}\}$,
$p(c_{1})=0.10$, $p(c_{2})=0.90$, and let $\hat{p}'$ and $\hat{p}''$
be two predicted distributions such that $\hat{p}'(c_{1})=0.05$,
$\hat{p}'(c_{2})=0.95$, $\hat{p}''(c_{1})=0.15$, and
$\hat{p}''(c_{2})=0.85$. If an EMQ $D$ satisfies \textbf{IMP} then
$D(p,\hat{p}')=D(p,\hat{p}'')$.

We contend that \textbf{IMP} is indeed a desirable property of any
EMQ, since underprediction and overprediction should be equally
penalized, unless there is a specific reason for not doing
so.\footnote{One might argue that underestimating the prevalence of a
class $c_{1}$ always implies overestimating the prevalence of another
class $c_{2}$. However, there are cases in which $c_{1}$ and $c_{2}$
are not equally important. For instance, if
$\mathcal{C}=\{c_{1},c_{2}\}$, with $c_{1}$ the class of patients that
suffer from a certain rare disease (say, one such that
$p(c_{1})=.0001$) and $c_{2}$ the class of patients who do not, the
class whose prevalence we really want to quantify is $c_{1}$, the
prevalence of $c_{2}$ being derivative. So, what we really care about
is that underestimating $p(c_{1})$ and overestimating $p(c_{1})$ are
equally penalized. The formulation of \textbf{IMP}, which involves
underestimation and overestimation in a perfectly symmetric way, is
strong enough that \textbf{IMP} is not satisfied (as we will see in
Section \ref{sec:measures}) by a number of important EMQs.}  If, in a
given application, we want to state that the two mistakes bring about
different costs, we should be able to explicitly state these costs as
parameters of the adopted measure.\footnote{In this case we enter the
realm of \emph{cost-sensitive quantification}, which is outside the
scope of this paper; see \cite[\S4\&\S5]{Forman:2008kx} and \cite[\S
10]{Gonzalez:2017it} for more on the relationships between
quantification and cost.} However, in the absence of any such explicit
statement, the two errors should be considered equally serious.

A further reason for insisting that an EMQ satisfies \textbf{IMP} is
that the parameters of a quantifier trained via supervised learning,
if optimized on a measure $D$ that penalizes (say) the underprediction
of $p(c)$ less than it penalizes its overprediction, will be such that
the quantifier will systematically tend to underpredict
$p(c)$. Depending on the type of parameters, this may be the result of
optimization carried out either implicitly (i.e., via supervised
learners that use $D$ as the loss to minimize -- see e.g.,
\citep{Esuli:2015gh}) or explicitly (i.e., via $k$-fold cross
validation).




So far we have discussed properties that, as we claim, should be
enjoyed by any EMQ. This is not the case for the next (and last) two
properties since they exclude each other (i.e., an EMQ may not enjoy
them both). We will claim that in some application contexts the former
is desirable while in other application contexts the latter is
desirable.
 
\begin{prop}
  \label{ax:rel}
  \textbf{Relativity} (\textbf{REL}).  For any codeframe
  $\mathcal{C}$, constant $a>0$, true distributions $p'$ and $p''$
  that only differ for the fact that, for two classes $c_{1}$ and
  $c_{2}$, $p'(c_{1})<p''(c_{1})$ and $p''(c_{1})<p''(c_{2})$ and
  $p''(c_{2})<p'(c_{2})$, if a predicted distribution $\hat{p}'$ that
  estimates $p'$ is such that $\hat{p}'(c_{1})=p'(c_{1})\pm a$ and
  $\hat{p}'(c_{2})=p'(c_{2})\mp a$, and a predicted distribution
  $\hat{p}''$ that estimates $p''$ is such that
  $\hat{p}''(c_{1})=p''(c_{1})\pm a$ and
  $\hat{p}''(c_{2})=p''(c_{2})\mp a$, and $\hat{p}'$ and $\hat{p}''$
  are identical except for the above, then it holds that
  $D(p',\hat{p}')>D(p'',\hat{p}'')$.  \hfill \mbox{} \qed
\end{prop}

\noindent In order to understand this fairly complex
formulation\footnote{The symbol $\pm$ stands for ``plus or minus''
while $\mp$ stands for ``minus or plus''; when symbol $\pm$ evaluates
to +, symbol $\mp$ evaluates to -, and vice versa.} let us see a
concrete example.

\begin{ex}
  Assume that $\mathcal{C}=\{c_{1},c_{2},c_{3},c_{4}\}$, and that
  $p',p'',\hat{p}',\hat{p}''$ are described by the following table:
  \begin{center}
    \begin{tabular}{|l||c|c|c|c|}
      \hline
      & $c_{1}$ & $c_{2}$ & $c_{3}$ & $c_{4}$ \\
      \hline
      $p'$        & 0.15 & 0.35 & 0.40 & 0.10 \\
      $\hat{p}'$  & 0.10 & 0.55 & 0.30 & 0.05 \\
      \hline
      $p''$       & 0.20 & 0.30 & 0.40 & 0.10 \\
      $\hat{p}''$ & 0.15 & 0.50 & 0.30 & 0.05 \\
      \hline
    \end{tabular}
  \end{center}
  \noindent This scenario is characterized by the fact that, of the
  only two classes ($c_{1}$ and $c_{2}$) that have different
  prevalence in $p'$ and $p''$, the one with the smallest true
  prevalence ($c_{1}$) in both $p'$ and $p''$ is underestimated by the
  same amount (0.05) by both $\hat{p}'$ and $\hat{p}''$. In this case
  $D$ penalizes (if it satisfies \textbf{REL}) $\hat{p}'$ more than it
  penalizes $\hat{p}''$, since $p'(c_{1})<p''(c_{1})$. \hfill \mbox{}
  \qed
\end{ex}

\noindent The rationale of \textbf{REL} is that an EMQ that satisfies
it, sanctions that an error of absolute magnitude $a$ is more serious
when the true class prevalence is smaller. \textbf{REL} may be a
desirable property in some applications of quantification.  Consider,
as an example, the case in which the prevalence $p(c)$ of pathology
$c$ as a cause of death in a population has to be estimated, for
epidemiological purposes, from verbal descriptions of the symptoms
that the deceased exhibited before dying \citep{King:2008fk}. In this
case, \textbf{REL} should arguably be a property of the EMQ; in fact,
predicting $\hat{p}'(c)=0.0101$ when $p'(c)=0.0001$ is a much more
serious mistake than predicting $\hat{p}''(c)=0.1100$ when
$p''(c)=0.1000$, since in the former case a very rare cause of death
is overestimated by two orders of magnitude (e.g., the presence of an
epidemic might mistakenly be inferred), while the same is not true in
the latter case.

However, in other applications of quantification \textbf{REL} may be
undesirable.  To see this, consider an example in which we want to
predict the prevalence $p(\textsf{NoShow})$ of the \textsf{NoShow}
class among the passengers booked on a flight with actual capacity $X$
(so that the airline can ``overbook'' additional
$\hat{p}(\textsf{NoShow})\cdot X$ seats). In this application,
relativity should arguably \emph{not} be a property of the evaluation
measure, since predicting $\hat{p}(\textsf{NoShow})=0.05$ when
$p(\textsf{NoShow})=0.10$ or predicting
$\hat{p}(\textsf{NoShow})=0.15$ when $p(\textsf{NoShow})=0.20$ brings
about the same cost to the airline (i.e., that $0.05\cdot X$ seats
will remain empty).
Applications such as this demand that the EMQ satisfies instead the
following property.

\begin{prop}
  \label{ax:abs}
  \textbf{Absoluteness} (\textbf{ABS}).  For any codeframe
  $\mathcal{C}$, constant $a>0$, true distributions $p'$ and $p''$
  that only differ for the fact that, for two classes $c_{1}$ and
  $c_{2}$, $p'(c_{1})<p''(c_{1})$ and $p''(c_{1})<p''(c_{2})$ and
  $p''(c_{2})<p'(c_{2})$, if a predicted distribution $\hat{p}'$ that
  estimates $p'$ is such that $\hat{p}'(c_{1})=p'(c_{1})\pm a$ and
  $\hat{p}'(c_{2})=p'(c_{2})\mp a$, and a predicted distribution
  $\hat{p}''$ that estimates $p''$ is such that
  $\hat{p}''(c_{1})=p''(c_{1})\pm a$ and
  $\hat{p}''(c_{2})=p''(c_{2})\mp a$, and $\hat{p}'$ and $\hat{p}''$
  are identical except for the above, then it holds that
  $D(p',\hat{p}')=D(p'',\hat{p}'')$.  \hfill \mbox{} \qed
\end{prop}

\noindent The formulation of \textbf{ABS} only differs from the
formulation of \textbf{REL} for its conclusion: while \textbf{REL}
stipulates that $D(p',\hat{p}')$ must be higher than
$D(p'',\hat{p}'')$, \textbf{ABS} states that the two must be
equal. The rationale of \textbf{ABS} is to guarantee that an error of
the same magnitude has the same impact on $D$ regardless of the true
prevalence of the class. \textbf{ABS} and \textbf{REL} are thus
mutually exclusive.

Note that \textbf{ABS} and \textbf{REL} are not redundant, i.e., they
do not cover the entire spectrum of possibilities (see Section
\ref{sec:Concordanceratio} for an example EMQ that enjoys
neither). For instance, an EMQ might consider an error more serious
when the true class prevalence is \emph{larger}, in which case it
would satisfy neither \textbf{REL} nor \textbf{ABS}. As the two
examples above show, there are applications that positively demand
\textbf{REL} to hold and others that positively demand
\textbf{ABS}. As a result, we will not claim that an EMQ must (or must
not) enjoy \textbf{REL} or \textbf{ABS}; we simply think it is
important to ascertain whether a given EMQ satisfies \textbf{REL} or
\textbf{ABS} or neither, since depending on this the EMQ may or may
not be adequate for the application one is tackling.
\fabscomment{Checked, July 5.}


\subsection{Reformulating \textbf{MON}, \textbf{IMP}, \textbf{REL},
\textbf{ABS}}
\label{sec:reformulating}
 
\noindent The formulations of four of the properties presented above
(namely, \textbf{MON}, \textbf{IMP}, \textbf{REL}, \textbf{ABS}) might
seem baroque, i.e., not as tight as they could be. In this section we
will try to simplify them, but for this we need to discuss a further
property.

Assume a codeframe $\mathcal{C}=\{c_{1}, ..., c_{n}\}$ partitioned
into $\mathcal{C}_{1}=\{c_{1}, ..., c_{k}\}$ and
$\mathcal{C}_{2}=\{c_{k+1}, ..., c_{n}\}$, and a true distribution $p$
on $\mathcal{C}$ such that $\sum_{c\in\mathcal{C}_{1}}p(c)=a$ for some
constant $0<a\leq 1$. We define the \emph{projection} of $p$ on
$\mathcal{C}_{1}$ as the distribution $p_{\mathcal{C}_{1}}$ on
$\mathcal{C}_{1}$ such that $p_{\mathcal{C}_{1}}(c)=\frac{p(c)}{a}$
for all $c\in\mathcal{C}_{1}$.

\begin{ex}
  Assume that $\mathcal{C}=\{c_{1},c_{2},c_{3},c_{4}\}$, that
  $\mathcal{C}_{1}=\{c_{1},c_{2},c_{3}\}$, and that $p$ is as in the
  1st row of the following table. The \emph{projection} of $p$ on
  $\mathcal{C}_{1}$ is then described in the 2nd row of the same
  table.
  \begin{center}
    \begin{tabular}{|l||c|c|c|c|}
      \hline
      & $c_{1}$ & $c_{2}$ & $c_{3}$ & $c_{4}$ \\
      \hline
      $p$                   & 0.32 & 0.00 & 0.48 & 0.20 \\
      $p_{\mathcal{C}_{1}}$ & 0.40 & 0.00 & 0.60 & --- \\
      \hline
    \end{tabular}
  \end{center} \hfill \mbox{} \qed
\end{ex}
\noindent Essentially, the projection on
$\mathcal{C}_{1}\subset\mathcal{C}$ of a distribution $p$ defined on
$\mathcal{C}$ is a distribution defined on $\mathcal{C}_{1}$ such that
the ratios between prevalences of classes that belong to
$\mathcal{C}_{1}$ are the same in $\mathcal{C}$ and $\mathcal{C}_{1}$.

We are now ready to describe Property \ref{ax:ind}.

\begin{prop}
  \label{ax:ind}
  \textbf{Independence} (\textbf{IND}). For any codeframes
  $\mathcal{C}=\{c_{1}$, ..., $c_{n}\}$,
  $\mathcal{C}_{1}=\{c_{1}, ..., c_{k}\}$ and
  $\mathcal{C}_{2}=\{c_{k+1}, ..., c_{n}\}$, for any true distribution
  $p$ on $\mathcal{C}$ and predicted distributions $\hat{p}'$ and
  $\hat{p}''$ on $\mathcal{C}$ such that $\hat{p}'(c)= \hat{p}''(c)$
  for all $c\in\mathcal{C}_{2}$, it holds that
  $D(p,\hat{p}')\leq D(p,\hat{p}'')$ if and only if
  $D(p_{\mathcal{C}_{1}},\hat{p}'_{\mathcal{C}_{1}})\leq
  D(p_{\mathcal{C}_{1}},\hat{p}''_{\mathcal{C}_{1}})$.  \hfill \mbox{}
  \qed
\end{prop}

\noindent If $D$ satisfies property \textbf{IND}, this essentially
means that when two predicted distributions estimate the prevalence of
all classes $\{c_{k+1}, ..., c_{n}\}$ identically, according to $D$
their relative merit is independent from these classes, and can thus
be established by focusing only on the remaining classes
$\{c_{1}, ..., c_{k}\}$.

We can now attempt to simplify the formulation of the \textbf{MON},
\textbf{IMP}, \textbf{REL}, \textbf{ABS} properties. For this
discussion we will take \textbf{MON} as an example, since similar
considerations also apply to the other three properties.

What we would like from a monotonicity property is to stipulate that
any even small increase in quantification error must generate an
increase in the value of $D(p,\hat{p})$. However, the notion of an
``increase in quantification error'' is non-trivial. To see this, note
that characterizing an increase in \emph{classification} error is
simple, since the units of classification (the unlabelled items) are
independent of each other: in a single-label context, to generate an
increase in classification error one just needs to switch the
predicted label of a single test items from correct to incorrect, and
the other items are not affected.\footnote{This is the basis of the
``Strict Monotonicity'' property discussed in
\citep{Sebastiani:2015zl} for the evaluation of classification
systems.} In a quantification context, instead, increasing the
difference between $p(c_{i})$ and $\hat{p}(c_{i})$ for some $c_{i}$
does not necessarily increase quantification error, since the
estimation(s) of some other class(es) in $\mathcal{C}/\{c_{i}\}$
is/are affected too, in many possible ways; in some cases the
quantification error across the entire codeframe $\mathcal{C}$
unequivocally increases, while in some other cases it is not clear
whether this happens or not, as the following example shows.

\begin{ex}
  \label{ex:multivariate}Assume that
  $\mathcal{C}=\{c_{1},c_{2},c_{3},c_{4}\}$, and assume the following
  true distribution $p$ and predicted distributions
  $\hat{p}',\hat{p}'',\hat{p}'''$:

  \begin{center}
    \begin{tabular}{|l||c|c|c|c|}
      \hline
      & $c_{1}$ & $c_{2}$ & $c_{3}$ & $c_{4}$ \\
      \hline
      \hline
      $p$          & 0.20 & 0.30 & 0.25 & 0.25 \\
      \hline
      $\hat{p}'$   & 0.25 & 0.15 & 0.30 & 0.30 \\
      $\hat{p}''$ & 0.35 & 0.15 & 0.25 & 0.25 \\
      $\hat{p}'''$  & 0.35 & 0.05 & 0.30 & 0.30 \\
      \hline
    \end{tabular}
  \end{center}

  \noindent In switching from $\hat{p}'$ to $\hat{p}''$ the
  quantification error on $c_{1}$ increases, but the quantification
  error on $c_{3}$ and $c_{4}$ decreases, so that it is not clear
  whether we should consider the quantification error on $\mathcal{C}$
  to increase or decrease. Conversely, in switching from $\hat{p}'$ to
  $\hat{p}'''$ the quantification errors on $c_{1}$ \emph{and} on the
  rest of the codeframe both increase.  
  \qed
\end{ex}

\noindent Example \ref{ex:multivariate} shows that the increase in the
quantification error on a single class says nothing about how the
quantification error on the entire codeframe varies.  As a result, in
\textbf{MON} we cannot stipulate (as we would have liked) that, in
switching from one predicted distribution to another, $D$ should
increase with the increase in the estimation error on a single class
$c_{1}$. The only thing we can do is to impose a monotonicity
condition on how $D$ behaves in a specific case, i.e., when the
increase in the estimation error on a class $c_{1}$ is exactly matched
by an estimation error (of identical magnitude but opposite sign) on
another class $c_{2}$ (which is what \textbf{MON} does) while the
estimation errors on all the other classes do not change.

The two predicted distributions $\hat{p}'$ and $\hat{p}''$ mentioned
in \textbf{MON} are such that
$\hat{p}'(c_{1})+\hat{p}'(c_{2})=\hat{p}''(c_{1})+\hat{p}''(c_{2})=a$
for some constant $0<a\leq 1$, while both
$\sum_{c\in\mathcal{C}/\{c_{1},c_{2}\}}\hat{p}'(c)$ and
$\sum_{c\in\mathcal{C}/\{c_{1},c_{2}\}}\hat{p}''(c)$ are equal to
$(1-a)$.  This means that, assuming that $D$ satisfies \textbf{IND},
we can reformulate \textbf{MON} in a way that disregards classes other
than $\{c_{1},c_{2}\}$ and considers instead the projection of $p$ on
$\{c_{1},c_{2}\}$. In other words, if $D$ satisfies \textbf{IND} we
can reformulate \textbf{MON}
in a way that tackles the problem in a binary quantification context
(instead of the more general single-label quantification context). The
fact that, in a binary context, $p(c_{2})=(1-p(c_{1}))$ for any (true
or predicted) distribution $p$, means that \textbf{MON} can be
reformulated by simply referring to just one of the two classes, i.e.,

\begin{prop}
  \label{ax:mon2}
  \textbf{Binary Strict Monotonicity} (\textbf{B-MON}).  For any
  codeframe $\mathcal{C}=\{c_{1},c_{2}\}$ and true distribution $p$,
  if predicted distributions $\hat{p}',\hat{p}''$ are such that
  $\hat{p}''(c_{1})<\hat{p}'(c_{1})\leq p(c_{1})$, then it holds that
  $D(p,\hat{p}')<D(p,\hat{p}'')$.  \hfill \mbox{} \qed
\end{prop}

\noindent As a result of what we have said in this section,
\textbf{B-MON} is, for any EMQ $D$ that satisfies \textbf{IND},
equivalent to \textbf{MON}. It is also much more compact since, among
other things, it makes reference to a single class
only. Considerations analogous to the ones above can be made for
\textbf{IMP}, \textbf{REL}, \textbf{ABS}. We reformulate them too as
below.
\begin{prop}
  \label{ax:imp2}
  \textbf{Binary Impartiality} (\textbf{B-IMP}).  For any codeframe
  $\mathcal{C}=\{c_{1},c_{2}\}$, true distribution $p$, predicted
  distributions $\hat{p}'$ and $\hat{p}''$, and constant $a\geq 0$
  such that $\hat{p}'(c_{1})=p(c_{1})+a$ and
  $\hat{p}''(c_{1})=p(c_{1})-a$, it holds that
  $D(p,\hat{p}')=D(p,\hat{p}'')$.
  \qed
\end{prop}
\begin{prop}
  \label{ax:rel2}
  \textbf{Binary Relativity} (\textbf{B-REL}).  For any codeframe
  $\mathcal{C}=\{c_{1},c_{2}\}$, constant $a>0$, true distributions
  $p'$ and $p''$ such that $p'(c_{1})<p''(c_{1})$ and
  $p''(c_{1})<p''(c_{2})$, if a predicted distribution $\hat{p}'$ that
  estimates $p'$ is such that $\hat{p}'(c_{1})=p'(c_{1})\pm a$ and a
  predicted distribution $\hat{p}''$ that estimates $p''$ is such that
  $\hat{p}''(c_{1})=p''(c_{1})\pm a$, then it holds that
  $D(p',\hat{p}')>D(p'',\hat{p}'')$.  \hfill \mbox{} \qed
\end{prop}
\begin{prop}
  \label{ax:abs2}
  \textbf{Binary Absoluteness} (\textbf{B-ABS}).  For any codeframe
  $\mathcal{C}=\{c_{1},c_{2}\}$, constant $a>0$, true distributions
  $p'$ and $p''$ such that $p'(c_{1})<p''(c_{1})$ and
  $p''(c_{1})<p''(c_{2})$, if a predicted distribution $\hat{p}'$ that
  estimates $p'$ is such that $\hat{p}'(c_{1})=p'(c_{1})\pm a$ and a
  predicted distribution $\hat{p}''$ that estimates $p''$ is such that
  $\hat{p}''(c_{1})=p''(c_{1})\pm a$, then it holds that
  $D(p',\hat{p}')=D(p'',\hat{p}'')$.  \hfill \mbox{} \qed
\end{prop}
\noindent In the next sections, instead of trying to prove that an EMQ
verifies Properties \ref{ax:mon}--\ref{ax:abs}, we will equivalently
(i) try to prove that it verifies \textbf{IND}, and if successful (ii)
try to prove that it verifies Properties \ref{ax:mon2}--\ref{ax:abs2};
the reason is, of course, the much higher simplicity and compactness
of the formulations of Properties \ref{ax:mon2}--\ref{ax:abs2} with
respect to Properties
\ref{ax:mon}--\ref{ax:abs}. \fabscomment{Checked, July 5.}

\section{Evaluation Measures for Single-Label Quantification}
\label{sec:measures}

\noindent In this section we turn to the functions that have been
proposed and used for evaluating quantification, and discuss whether
they comply or not with the properties that we have discussed in
Section \ref{sec:propertiesforSLQ}. In many cases these functions were
originally proposed for evaluating the binary case; since the
extension to SLQ is usually straightforward, for each EMQ we indicate
its original proponent or user (on this see also Table
\ref{sec:adopters}) and disregard whether it was originally used just
for BQ or for the full-blown SLQ.

We will discuss 9 measures proposed as EMQs in the literature, and for
each of them we will be interested in whether they satisfy or not
Properties \ref{ax:ioi} to \ref{ax:ind}. Giving $9\times8=72$ proofs
in detail would make the paper excessively long and boring: as a
result, only some of these proofs will be given in detail, while for
others we will only give hints at how they can be easily obtained via
the same lines of reasoning used in other cases.  In several cases,
given a measure $D$ and a property $\pi$, one can simply show that $D$
does \emph{not} enjoy $\pi$ via a counterexample. Since the same
scenario can serve as a counterexample for showing that $\pi$ is not
enjoyed by several measures, we formulate each such scenario in the
form of a table that shows which measures the scenario rules out. In
the appendix we include a table each for properties \textbf{MAX}
(Appendix \ref{sec:testingforMAX}), \textbf{IMP} (Appendix
\ref{sec:testingforIMP}), \textbf{REL} (Appendix
\ref{sec:testingforREL}), \textbf{ABS} (Appendix
\ref{sec:testingforABS}); in this section, when discussing the
property in the context of a specific measure that does not enjoy it,
we will simply refer the reader to the appropriate
table. \fabscomment{Checked, August 31.}


\subsection{Absolute Error}
\label{sec:ae}

\begin{figure}[htbp]
  \begin{center}
    \includegraphics[width=.32\textwidth]{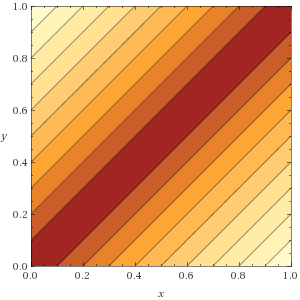}
    \includegraphics[width=.32\textwidth]{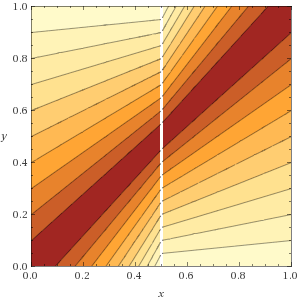}
    \includegraphics[width=.32\textwidth]{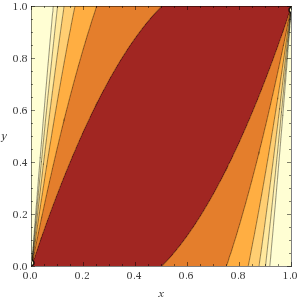} \\
    $\aae$ \hspace{10.5em} $\nae$ \hspace{10.5em} $\rae$ \\
    \bigskip
    \includegraphics[width=.32\textwidth]{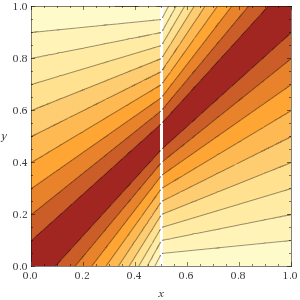}
    \includegraphics[width=.32\textwidth]{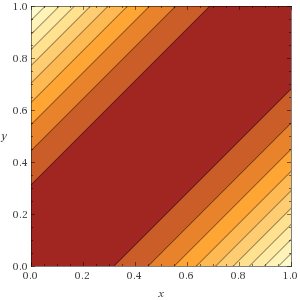}
    \includegraphics[width=.32\textwidth]{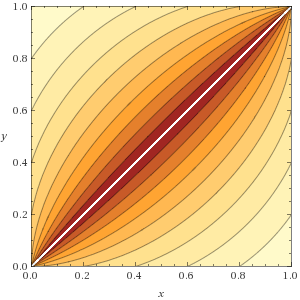} \\
    $\nrae$ \hspace{10.5em} $\se$ \hspace{10.5em} $\dr$ \\
    \bigskip
    \includegraphics[width=.32\textwidth]{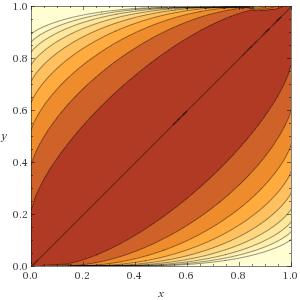}
    \includegraphics[width=.32\textwidth]{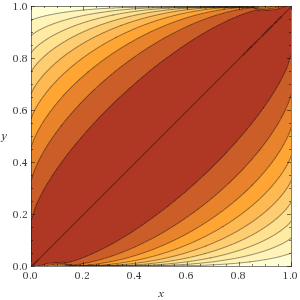}
    \includegraphics[width=.32\textwidth]{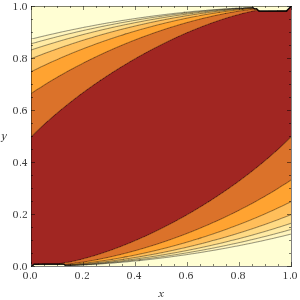} \\
    $\kld$ \hspace{10.5em} $\nkld$ \hspace{10.5em} $\pd$ \\
    \caption{\blue{2D plots (for a binary quantification task) for the
    nine EMQs of Tables \ref{sec:propertiesofEMQs} and
    \ref{sec:adopters}; $p(c_{1})$ and $p(c_{2})$ are represented as
    $x$ and $(1-x)$, respectively, while $\hat{p}(c_{1})$ and
    $\hat{p}(c_{2})$ are represented as $y$ and $(1-y)$. Darker areas
    represent values closer to 0 (i.e., smaller error) while lighter
    areas represent values more distant from 0 (i.e., higher error).}
    \fabscomment{Questi grafi li dovremmo ricontrollare: corrispondono
    alle nostre intuizioni? Un plot simmetrico rispetto alla diagonale
    ((0,0),(1,1)) dovrebbe indicare una EMD che gode di \textbf{IMP}
    (R: NO.); ma come mai \textbf{RAE} esibisce un plot tutt'altro che
    simmetrico? }}
    \label{fig:2d}
  \end{center}
\end{figure}

\begin{figure}[htbp]
  \begin{center}
    \includegraphics[width=.32\textwidth]{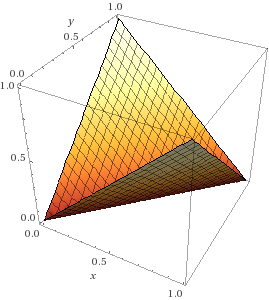}
    \includegraphics[width=.32\textwidth]{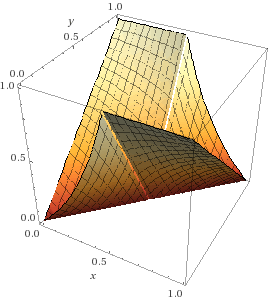}
    \includegraphics[width=.32\textwidth]{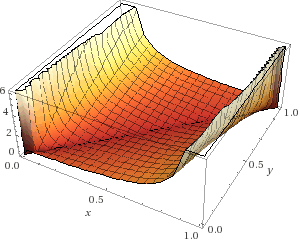} \\
    $\aae$ \hspace{10.5em} $\nae$ \hspace{10.5em} $\rae$ \\
    \bigskip
    \includegraphics[width=.32\textwidth]{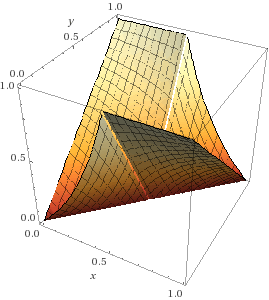}
    \includegraphics[width=.32\textwidth]{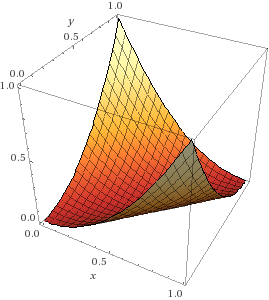}
    \includegraphics[width=.32\textwidth]{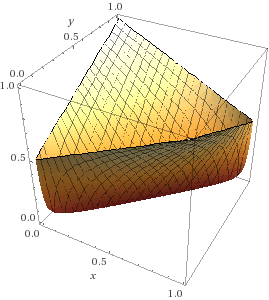} \\
    $\nrae$ \hspace{10.5em} $\se$ \hspace{10.5em} $\dr$ \\
    \bigskip
    \includegraphics[width=.32\textwidth]{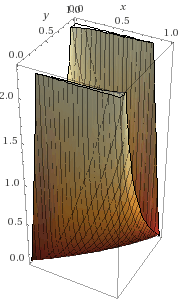}
    \includegraphics[width=.32\textwidth]{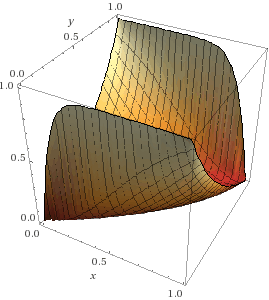}
    \includegraphics[width=.32\textwidth]{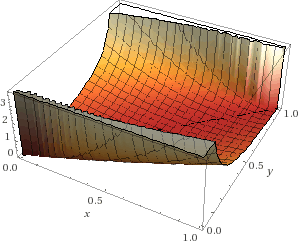} \\
    $\kld$ \hspace{10.5em} $\nkld$ \hspace{10.5em} $\pd$ \\
    \caption{\blue{3D plots (for a binary quantification task) for the
    nine EMQs of Tables \ref{sec:propertiesofEMQs} and
    \ref{sec:adopters}; $p(c_{1})$ and $p(c_{2})$ are represented as
    $x$ and $(1-x)$, respectively, while $\hat{p}(c_{1})$ and
    $\hat{p}(c_{2})$ are represented as $y$ and $(1-y)$; error is
    represented as $z$ (higher values of $z$ represent higher error).}
    \fabscomment{Questi grafi li dovremmo ricontrollare: corrispondono
    alle nostre intuizioni?}}
    \label{fig:3d}
  \end{center}
\end{figure}

\noindent The simplest EMQ is \emph{Absolute Error} ($\aae$), which
corresponds to the average (across the classes in $\mathcal{C}$)
absolute difference between the predicted class prevalence and the
true class prevalence; i.e.,
\begin{equation}
  \label{eq:AE}
  \aae(p,\hat{p})=\frac{1}{|\mathcal{C}|}\sum_{c\in \mathcal{C}}|\hat{p}(c)-p(c)|
\end{equation}
\noindent \blue{A 2D plot of $\aae$ (and of the other 8
measures we will discuss next) for the case of binary quantification
is displayed in Figure \ref{fig:2d}; Figure \ref{fig:3d} displays the
same plots in 3D.}


It is easy to prove that $\aae$ enjoys \textbf{IoI}, \textbf{NN},
\textbf{MON}, \textbf{IMP}, \textbf{ABS}, \textbf{IND}. While some of
these proofs are trivial, we report them in detail (in Appendix
\ref{sec:ThePropertiesOfAE}) in order to show how the same arguments
can be used to prove the same for many of the EMQs to be discussed
later in this section.

Instead, as shown in Appendix \ref{sec:testingforMAX}, $\aae$ does not
enjoy \textbf{MAX}, because its range depends on the true distribution
$p$. More specifically, $\aae$ ranges between 0 (best) and
\begin{equation}
  \label{eq:zae}
  z_{\aae}=\displaystyle\frac{2(1-\displaystyle\min_{c\in
  \mathcal{C}}p(c))}{|\mathcal{C}|}
\end{equation}
\noindent (worst), i.e., its range depends also on the cardinality of
$\mathcal{C}$.  In fact, it is easy to verify that, given a true
distribution $p$ on $\mathcal{C}$, the perverse estimator of $p$ is
the one such that (a) $\hat{p}(c^{*})=1$ for class
$c^{*}=\arg\min_{c\in\mathcal{C}}p(c)$, and (b) $\hat{p}(c)=0$ for all
$c\in \mathcal{C}/\{c^{*}\}$.  In this case, the \emph{total} absolute
error derives (i) from overestimating $p(c^{*})$, which brings about
an error of $(1-p(c^{*}))$, and (ii) from underestimating $p(c)$ for
all $c\in \mathcal{C}/\{c^{*}\}$, which collectively brings about an
additional error of $(1-p(c^{*}))$. The \emph{mean} absolute error is
obtained by dividing this $2(1-p(c^{*}))$ quantity by $|\mathcal{C}|$.

Concerning \textbf{REL}, just note that since $\aae$ satisfies
\textbf{ABS}, it cannot (as observed in Section
\ref{sec:propertiesforSLQ}) satisfy \textbf{REL}. (That $\aae$ does
not enjoy \textbf{REL} is also shown via a counterexample in Appendix
\ref{sec:testingforREL}.)

The properties that $\aae$ enjoys (and those it does not enjoy) are
conveniently summarized in Table \ref{sec:propertiesofEMQs}, along
with the same for all the measures discussed in the rest of this
paper.


In the literature, $\aae$ also goes by the name of \emph{Variational
Distance} \cite[\S 4]{Csiszar:2004fk},\citep{Lin:1991vn,Zhang:2010kx},
or \emph{Percentage Discrepancy}
\citep{Esuli:2010fk,Baccianella:2013fg}. Also, if viewed as a generic
function of dissimilarity between vectors (and not just probability
distributions), $\aae$ is nothing else than the well-known
``city-block distance'' normalized by the number of classes. Some
recent papers \citep{Beijbom:2015yg,Gonzalez:2017ff} that tackle
quantification in the context of ecological modelling discuss or use,
as an EMQ, \emph{Bray-Curtis dissimilarity} (BCD), a measure popular
in ecology for measuring the dissimilarity of two samples. However,
when used to measure the dissimilarity of two probability
distributions, BCD defaults to $\aae$; as a result we will not analyse
BCD any further.

Note that $\aae$ often goes by the name of \emph{Mean} Average Error;
for simplicity, for this and the other measures we discuss in the rest
of this paper we will omit the qualification ``Mean'', since every
measure mediates across the class-specific values in its own way.

As an EMQ, $\aae$ was used for the first time by
\citep{Saerens:2002uq}, and in many other papers ever since.  For
$\aae$ and for all the other EMQs discussed in this paper, Table
\ref{sec:adopters} lists the papers where the measure has been
proposed and those which have subsequently used it for evaluation
purposes.  \fabscomment{Checked, Aug 31.}



\subsection{Normalized Absolute Error}
\label{sec:nae}

\noindent Following what we have said in Section \ref{sec:ae}, a
normalized version of $\aae$ that always ranges between 0 (best) and 1
(worst) can be obtained as
\begin{equation}
  \label{eq:NAE}
  \begin{aligned}
    \nae(p,\hat{p}) & = \ \dfrac{\aae(p,\hat{p})}{z_{\aae}}
    = \ \frac{\sum_{c\in
    \mathcal{C}}|\hat{p}(c)-p(c)|}{2(1-\displaystyle\min_{c\in
    \mathcal{C}}p(c))}
  \end{aligned}
\end{equation}
\noindent where $z_{\aae}$ is as in Equation \ref{eq:zae}.
It is easy to verify that $\nae$ enjoys \textbf{IoI}, \textbf{NN},
\textbf{MON}, \textbf{IMP}, \textbf{IND}. $\nae$ also enjoys (by
construction) \textbf{MAX}.

Given that $\nae$ is just a normalized version of $\aae$, and given
that $\aae$ enjoys \textbf{ABS}, one might expect that $\nae$ enjoys
\textbf{ABS} too. Surprisingly enough, this is not the case, as shown
in the counterexample of Appendix \ref{sec:testingforABS}. The reason
for this is that, for the two distributions $p'$ and $p''$ (and their
respective predicted distributions $\hat{p}'$ and $\hat{p}''$)
mentioned in the formulation of Property \ref{ax:abs} (\textbf{ABS}),
and exemplified in the counterexample of Appendix
\ref{sec:testingforABS}, the numerator of Equation \ref{eq:NAE} is the
same but the denominator (i.e., the normalizing constant) is
different, which means that the value of $\nae$ is also different.
$\nae$ does not enjoy \textbf{REL} either, as also shown in Appendix
\ref{sec:testingforREL}).

$\nae$ was discussed for the first time by \cite{Esuli:2014uq}. With a
similar intent, in a binary quantification context
\cite{Barranquero:2015fr} proposed \emph{Normalized Absolute Score}
($\nas$). $\nas$ is an accuracy (and not an error) measure; when
viewed as an error measure, it is defined as
\begin{equation}
  \label{eq:NAS}
  \begin{aligned}
    \nas(p,\hat{p})= \frac{|p(c)-\hat{p}(c)|}{\max\{p(c),(1-p(c))\}}
  \end{aligned}
\end{equation}
\noindent where $c$ is any class in $\mathcal{C}=\{c_{1},c_{2}\}$. We
will not discuss $\nas$ in detail since (a) it is only defined for the
binary case, and (b) it is easy to show that in this case it coincides
with $\nae$. \fabscomment{Checked, Sep 1.}


\subsection{Relative Absolute Error}
\label{sec:rae}

\noindent \emph{Relative Absolute Error} ($\rae$) relativises the
value $|\hat{p}(c)-p(c)|$ in Equation \ref{eq:AE} to the true class
prevalence, i.e.,
\begin{equation}
  \label{eq:rae}
  \rae(p,\hat{p})=\frac{1}{|\mathcal{C}|}\sum_{c\in 
  \mathcal{C}}\displaystyle\frac{|\hat{p}(c)-p(c)|}{p(c)}
\end{equation}
\noindent $\rae$ may be undefined in some cases, due to the presence
of zero denominators. To solve this problem, in computing $\rae$ we
can smooth both $p(c)$ and $\hat{p}(c)$ via additive smoothing, i.e.,
we take
\begin{equation}
  \label{eq:smoothing}
  p_{s}(c)=\frac{\epsilon+p(c)}{\epsilon|\mathcal{C}|+\displaystyle\sum_{c\in 
  \mathcal{C}}p(c)}
\end{equation}
\noindent where $p_{s}(c)$ denotes the smoothed version of $p(c)$ and
the denominator is just a normalizing factor (same for the
$\hat{p}_{s}(c)$'s); the quantity $\epsilon=\frac{1}{2 |\sigma|}$ is
often used \blue{(and will always be used in the rest of this paper)}
as a smoothing factor. The smoothed versions of $p(c)$ and
$\hat{p}(c)$ are then used in place of their original non-smoothed
versions in Equation \ref{eq:rae}; as a result, $\rae$ is always
defined.

Using arguments analogous to the ones used for $\aae$ in Appendix
\ref{sec:ThePropertiesOfAE}, it is immediate to show that $\rae$
enjoys \textbf{IoI}, \textbf{NN}, \textbf{MON}, \textbf{IMP},
\textbf{IND}. It also enjoys \textbf{REL} by construction, which means
that it does not enjoy \textbf{ABS}. Analogously to $\aae$, the fact
$\rae$ does not enjoy \textbf{MAX}, as shown via the counterexample in
Appendix \ref{sec:testingforMAX}.


It is easy to show that $\rae$ ranges between 0 (best) and
\begin{equation}
  \label{eq:zrae}
  z_{\rae}=\displaystyle\frac{|\mathcal{C}|-1+\displaystyle\frac{1-
  \displaystyle\min_{c\in \mathcal{C}}p(c)}{\displaystyle\min_{c\in 
  \mathcal{C}}p(c)}}{|\mathcal{C}|}
\end{equation}
\noindent (worst), i.e., its range depends also on the cardinality of
$\mathcal{C}$. In fact, similarly to the case of $\aae$, it is easy to
verify that, given a true distribution $p$ on $\mathcal{C}$, the
perverse estimator of $p$ is obtained when (a) $\hat{p}(c)=1$ for the
class $c^{*}=\arg\min_{c\in\mathcal{C}}p(c)$, and (b) $\hat{p}(c)=0$
for all $c\in \mathcal{C}/\{c^{*}\}$. In this case, the \emph{total}
relative absolute error derives (i) from overestimating $p(c^{*})$,
which brings about an error of $\frac{1-p(c^{*})}{p(c^{*})}$, and (ii)
from underestimating $p(c)$ for all $c\in \mathcal{C}/\{c^{*}\}$,
which brings about an additional error of 1 for each class in
$\mathcal{C}/\{c^{*}\}$. The value of $\rae$ is then obtained by
dividing the resulting
$(|\mathcal{C}|-1+\displaystyle\frac{1- p(c^{*})}{p(c^{*})})$ by
$|\mathcal{C}|$.

As an EMQ, $\rae$ was used for the first time by
\cite{Gonzalez-Castro:2010fk}, and by several other papers after
it. \fabscomment{Checked, Aug 31.}


\subsection{Normalized Relative Absolute Error}
\label{sec:nmrae}

\noindent Following what we have said in Section \ref{sec:rae}, a
normalized version of $\rae$ that always ranges between 0 (best) and 1
(worst) can thus be obtained as
\begin{equation}
  \label{eq:nrae}
  \nrae(p,\hat{p})= \frac{\rae(p,\hat{p})}{z_{\rae}}= \frac{\displaystyle\sum_{c\in 
  \mathcal{C}}\displaystyle\frac{|\hat{p}(c) - 
  p(c)|}{p(c)}}{|\mathcal{C}|-1+\displaystyle\frac{1-
  \displaystyle\min_{c\in \mathcal{C}}p(c)}{\displaystyle\min_{c\in 
  \mathcal{C}}p(c)}}
\end{equation}
\noindent where $z_{\rae}$ is as in Equation \ref{eq:zrae}. Since the
various denominators of Equation \ref{eq:nrae} may be undefined, the
smoothed values of Equation \ref{eq:smoothing} must be used in
Equation \ref{eq:nrae} too.

It is straightforward to verify that $\nrae$, which was first proposed
by \cite{Esuli:2014uq}, enjoys \textbf{IoI}, \textbf{NN},
\textbf{MON}, \textbf{IMP}, \textbf{IND}, and also enjoys (by
construction) \textbf{MAX}.

Somehow similarly to what we said in Section \ref{sec:nae} about
$\nae$ and \textbf{ABS}, given that $\nrae$ is just a normalized
version of $\rae$, and given that $\rae$ enjoys \textbf{REL}, one
might expect that $\nrae$ enjoys \textbf{REL} too. Again, this is not
the case, as shown in the counterexample of Appendix
\ref{sec:testingforREL}. The reason for this is that, for the two
distributions $p'$ and $p''$ (and their respective predicted
distributions $\hat{p}'$ and $\hat{p}''$) mentioned in the formulation
of Property \ref{ax:rel} (\textbf{REL}), and exemplified in the
counterexample of Appendix \ref{sec:testingforREL}, while $\rae$ (the
numerator of Equation \ref{eq:nrae}) does enjoy \textbf{REL}, the
normalizing constant (the denominator of Equation \ref{eq:nrae})
invalidates it, since it is different for $p'$ and $p''$. $\nae$ does
not enjoy \textbf{ABS} either, as also shown in Appendix
\ref{sec:testingforABS}).  \fabscomment{Checked, Sep 1.}



\subsection{Squared Error}
\label{sec:mse}

\noindent Another measure that has been used in the quantification
literature
is \emph{Squared Error} ($\se$), defined as
\begin{equation}
  \begin{aligned}
    \label{eq:L2}
    \se(p,\hat{p}) = \frac{1}{|\mathcal{C}|}\sum_{c\in \mathcal{C}}
    (p(c)-\hat{p}(c))^{2}
  \end{aligned}
\end{equation}
\noindent When viewed as a generic function of dissimilarity between
vectors (and not just probability distributions), $\se$ is the
well-known $L^{2}$-distance. As an EMQ, $\se$ was used for the first
time by \cite{Bella:2010kx}.

The mathematical form of $\se$ is very similar to that of $\aae$, and
it can be trivially shown that $\se$ enjoys all the properties that
$\aae$ enjoys and does not enjoy all the properties that $\aae$ does
not enjoy. In particular, $\se$ does not enjoy \textbf{MAX} since
$\se$ ranges between 0 (best) and
\begin{equation}
  \label{eq:zse}
  z_{\se}=\displaystyle\frac{(1-p(c^{*}))^{2}+\sum_{c\in\mathcal{C}/\{c^{*}\}}p(c)^{2}}{|\mathcal{C}|}
\end{equation}
\noindent (worst), where $c^{*}=\arg\min_{c\in\mathcal{C}}p(c)$; i.e.,
the range of $\se$ depends on $p$ and $|\mathcal{C}|$. In fact,
similarly to the case of $\aae$, it is easy to verify that the
perverse estimator of a true distribution $p$ is the one such (a)
$\hat{p}(c^{*})=1$ and (b) $\hat{p}(c)=0$ for all
$c\in \mathcal{C}/\{c^{*}\}$. In this case, the squared error derives
(i) from overestimating $p(c^{*})$, which brings about an error of
$\frac{(1-p(c^{*}))^{2}}{|\mathcal{C}|}$, and (ii) from
underestimating $p(c)$ for all $c\in \mathcal{C}/\{c^{*}\}$, which
brings about an additional error of $\frac{p(c)^{2}}{|\mathcal{C}|}$
for each class in $\mathcal{C}/\{c^{*}\}$. We could thus define a
normalized version of $\se$ as
\begin{equation}
  \label{eq:nse}
  \nse(p,\hat{p})= 
  \frac{\se(p,\hat{p})}{z_{\se}}= \frac{\sum_{c\in \mathcal{C}}
  (p(c)-\hat{p}(c))^{2}}{(1-p(c^{*}))^{2}+\sum_{c\in\mathcal{C}/\{c^{*}\}}p(c)^{2}}
\end{equation}
\noindent which would, quite obviously, enjoy and not enjoy exactly
the same properties that $\nae$ enjoys and does not enjoy.

\blue{$\se$ is structurally similar to $\aae$ but (as can also
be appreciated from Figure \ref{fig:2d}) is less sensitive than it,
i.e., it is always the case that $\se(p,\hat{p})\leq \aae(p,\hat{p})$
(since it is always the case that
$(p(c)-\hat{p}(c))^{2}\leq |p(c)-\hat{p}(c)|$).}

In the binary quantification literature, other proxies of $\se$ have
been used; one example is \emph{Normalized Squared Score}
\citep{Barranquero:2015fr}, defined as
$\nss(p(c),\hat{p}(c))\equiv
1-(\frac{p(c)-\hat{p}(c)}{\max\{p(c),(1-p(c))\}})^{2}$, where $c$ is
any class in $\mathcal{C}=\{c_{1},c_{2}\}$. Similarly to the $\nas$
measure discussed at the end of Section \ref{sec:ae}, $\textrm{NSS}$
is an accuracy (and not an error) measure; when viewed as an error
measure, it would be defined as
\begin{equation}
  \label{eq:NSS}
  \begin{aligned}
    \nss(p,\hat{p})=
    (\frac{p(c)-\hat{p}(c)}{\max\{p(c),(1-p(c))\}})^{2}
  \end{aligned}
\end{equation}
\noindent where $c$ is any class in $\mathcal{C}=\{c_{1},c_{2}\}$. We
will not discuss $\nss$ in detail since (a) it is only defined for the
binary case, and (b) it is easy to show that in this case it coincides
with $\nse$. \fabscomment{Checked, Sep 1.}


\subsection{Discordance Ratio}
\label{sec:Concordanceratio}

\noindent \cite{Levin:2017dq} introduce an EMQ that they call
\emph{Concordance Ratio} (CR).  $\cra$ is a measure of accuracy, and
not a measure of error; for better consistency with the rest of this
paper, instead of $\cra$ we consider what might be called
\emph{Discordance Ratio}, i.e., its complement $\dr=(1-\cra)$, defined
as
\begin{equation}
  \begin{aligned}
    \label{eq:discordanceratio}
    \dr(p,\hat{p})= & \ 1 - \cra \\
    = & \ 1-\frac{1}{|\mathcal{C}|}\sum_{c\in\mathcal{C}}\dfrac{\min\{(p(c),\hat{p}(c)\}}{\max\{(p(c),\hat{p}(c)\}} \\
    = & \ \frac{1}{|\mathcal{C}|}\sum_{c\in\mathcal{C}}\dfrac{\max\{(p(c),\hat{p}(c)\}-\min\{(p(c),\hat{p}(c)\}}{\max\{(p(c),\hat{p}(c)\}} \\
    = & \
    \frac{1}{|\mathcal{C}|}\sum_{c\in\mathcal{C}}\dfrac{|(p(c)-\hat{p}(c)|}{\max\{(p(c),\hat{p}(c)\}}
  \end{aligned}
\end{equation}
\noindent $\dr$ is undefined when, for a given class $c$, both $p(c)$
and $\hat{p}(c)$ are zero; the smoothed values of Equation
\ref{eq:smoothing} must thus be used within Equation
\ref{eq:discordanceratio} in order to avoid this problem.

It is easy to verify, along the lines sketched in Appendix
\ref{sec:ThePropertiesOfAE}, that $\dr$ enjoys \textbf{IoI},
\textbf{NN}, \textbf{MON}, \textbf{IND}. $\dr$ also enjoys
\textbf{REL}; this can be seen by the fact that, for the same amount
$a$ of misprediction,
$\sum_{c\in\mathcal{C}}\frac{\min\{(p(c),\hat{p}(c)\}}{\max\{(p(c),\hat{p}(c)\}}$
is smaller (hence $\dr(p,\hat{p})$ is larger) when the true prevalence
of the class $c_{1}$ mentioned in the formulation of Property
\ref{ax:rel} (\textbf{REL}) is smaller. Instead, $\dr$ enjoys neither
\textbf{MAX}, nor \textbf{IMP}, nor \textbf{ABS}, as shown in
Appendixes \ref{sec:testingforMAX}, \ref{sec:testingforIMP} and
\ref{sec:testingforABS}, respectively.  \fabscomment{Checked, Sep 1.}


\subsection{Kullback-Leibler Divergence}
\label{sec:kld}

\noindent An EMQ that has become somehow standard in the evaluation of
single-label (and, \textit{a fortiori}, binary) quantification, is
\emph{Kullback-Leibler Divergence} ($\kld$ -- also called
\emph{Information Divergence}, or \emph{Relative Entropy})
\citep{Csiszar:2004fk} and defined as\footnote{In Equation
\ref{eq:kld} and in the rest of this paper the $\log$ operator denotes
the natural logarithm.}
\begin{equation}
  \label{eq:kld}
  \kld(p,\hat{p}) = \sum_{c\in \mathcal{C}} 
  p(c)\log\frac{p(c)}{\hat{p}(c)}
\end{equation}
\noindent
As an EMQ, $\kld$ was used for the first time (under the name
\emph{Normalized Cross-Entropy}) by \cite{Forman:2005fk}. It should
also be noted that $\kld$ has been adopted as the official evaluation
measure of the only quantification-related shared task that has been
organized so far, Subtask D ``Tweet Quantification on a 2-point
Scale'' of SemEval-2016 and Semeval-2017 ``Task 4: Sentiment Analysis
in Twitter'' \citep{Nakov:2016ty,Nakov:2017ty}.

$\kld$ may be undefined in some cases. While the case in which
$p(c)=0$ is not problematic (since continuity arguments indicate that
$0 \log \frac{0}{a}$ should be taken to be 0 for any $a\geq 0$), the
case in which $\hat{p}(c)=0$ and $p(c)>0$ is indeed problematic, since
$a\log\frac{a}{0}$ is undefined for $a>0$.  To solve this problem, we
smooth values in the same way as already described in Section
\ref{sec:rae}; as a result, $\kld$ is always defined.

That $\kld$ enjoys \textbf{IoI} and \textbf{NN} is not immediately
evident (since $p(c)\log\frac{p(c)}{\hat{p}(c)}$ is negative whenever
$p(c)<\hat{p}(c)$), but has been proven before \cite[p.\
423]{Csiszar:2004fk}. Indeed, $\kld$ is a well-known member of the
class of \emph{f-divergences} \citep{Ali:1966lq} \citep[\S
4]{Csiszar:2004fk}, a class of functions that measure the difference
between two probability distributions, and that all enjoy \textbf{IoI}
and \textbf{NN}.

The fact that $\kld$ enjoys \textbf{MON} is also not self-evident,
essentially for the same reasons for which it is not self-evident that
it enjoys \textbf{IoI} and \textbf{NN}. The proof that $\kld$ enjoys
\textbf{MON} is given in Appendix \ref{sec:provingmon}, where we use
the fact that $\kld$ enjoys \textbf{IND} (something which can be
easily shown via the arguments used in Appendix
\ref{sec:ThePropertiesOfAE}) and thus limit ourselves to proving that
it enjoys \textbf{B-MON}.

The fact that $\kld$ enjoys neither \textbf{MAX}, nor \textbf{IMP},
nor \textbf{REL}, nor \textbf{ABS} is shown in Appendixes
\ref{sec:testingforMAX}, \ref{sec:testingforIMP},
\ref{sec:testingforREL}, \ref{sec:testingforABS},
respectively. Concerning \textbf{MAX} we note that, in theory, the
upper bound of $\kld$ is not finite, since Equation \ref{eq:kld} has
predicted probabilities, and not true probabilities, at the
denominator. That is, by making a predicted probability $\hat{p}(c)$
infinitely small we can make $\kld$ infinitely large.  However, since
we use smoothed values, the fact that both $p$ and $\hat{p}$ are
lower-bounded by $\epsilon$, and not by 0, has the consequence that
$\kld$ has a finite upper bound.  The perverse estimator for $\kld$ is
the one such (a) $\hat{p}(c^{*})=1$ and (b) $\hat{p}(c)=0$ for all
$c\in \mathcal{C}/\{c^{*}\}$. The value of this estimator is
\begin{equation}
  \label{eq:kldupperbound}
  z_{\kld}(p,\hat{p})=p_{s}(c^{*})\log\frac{p_{s}(c^{*})}{1-(|\mathcal{C}|-1)\cdot \epsilon}+\sum_{c\in\mathcal{C}/\{c^{*}\}}p_{s}(c)\log\frac{p_{s}(c)}{\epsilon}
\end{equation}
\noindent which shows that the range of $\kld$ depends on $p$, the
cardinality of $\mathcal{C}$, and even on the value of
$\epsilon$. This is a further proof that $\kld$ does not enjoy
\textbf{MAX}. \fabscomment{Checked, Sep 1.}





\subsection{Normalized Kullback-Leibler Divergence}
\label{sec:nkld}

\noindent Given what we have said in Section \ref{sec:kld}, one might
define a normalized version of $\kld$ (i.e., one that also enjoys
\textbf{MAX}) as $\frac{\kld(p,\hat{p})}{z_{\kld}(p,\hat{p})}$, where
$z_{\kld}(p,\hat{p})$ is as in Equation
\ref{eq:kldupperbound}. \cite{Esuli:2014uq} follow instead a different
route, and define a normalized version of $\kld$ by applying to it a
logistic function,\footnote{Since the standard logistic function
$\frac{e^{x}}{e^{x}+1}$ ranges (for the domain $[0,+\infty)$ we are
interested in) on [$\frac{1}{2}$,1], a multiplication by 2 is applied
in order for it to range on [1,2], and 1 is subtracted in order for it
to range on [0,1], as desired.} i.e.,\footnote{\cite{Esuli:2014uq}
mistakenly defined $\nkld(p,\hat{p})$ as
$\frac{e^{\kld(p,\hat{p})}-1}{e^{\kld(p,\hat{p})}}$; this was later
corrected into the formulation of Equation \ref{eq:nkld} (which is
equivalent to $\frac{e^{\kld(p,\hat{p})}-1}{e^{\kld(p,\hat{p})}+1}$)
by \cite{Gao:2016uq}.}
\begin{equation} \label{eq:nkld}
  \begin{aligned}
    \nkld(p,\hat{p}) & = 2\frac{e^{\kld(p,\hat{p})}}{e^{\kld(p,\hat{p})}+1}-1 \\
  \end{aligned}
\end{equation}
\noindent Like other previously discussed measures, also $\nkld$ may
be undefined in some cases; therefore, also in computing $\nkld$ we
need to use the smoothed values of Equation \ref{eq:smoothing} in
place of the original $p(c)$'s and $\hat{p}(c)$'s.

$\nkld$ enjoys some of our properties of interest for the simple
reason that $\kld$ enjoys them; it is easy to verify that this is the
case of \textbf{IoI} and \textbf{NN}.  $\nkld$ also enjoys
\textbf{MON} and \textbf{IND}; this descends from the fact that
$\nkld(d,d')<\nkld(d,d'')$ if and only if $\kld(d,d')<\kld(d,d'')$
(this derives from the fact that the logistic function is a monotonic
transformation) and from the fact that $\kld$ enjoys \textbf{MON} and
\textbf{IND}, respectively. Concerning \textbf{MAX}, $\nkld$ enjoys it
by construction, because when a predicted prevalence $\hat{p}(c)$
tends to 0 $\kld$ tends to $+\infty$, and $\nkld$ thus tends to
1.\footnote{\label{foot:nkld}This is true only at a first
approximation, though. In more precise terms, the maximum value that
$\nkld$ can have is strictly smaller than 0 because the maximum value
that $\kld$ can have is finite (see Equation \ref{eq:kldupperbound})
and, as discussed at the end of Section \ref{sec:kld}, dependent on
$p$, on the cardinality of $\mathcal{C}$, and even on the value of
$\epsilon$; as a result, the maximum value that $\nkld$ can have is
also dependent on these three variables (although it is always very
close to 1 -- see the example in Appendix \ref{sec:testingforMAX}).}


The fact that $\kld$ enjoys neither \textbf{IMP}, nor \textbf{REL},
nor \textbf{ABS}, is shown in Appendixes \ref{sec:testingforIMP},
\ref{sec:testingforREL}, \ref{sec:testingforABS},
respectively. \fabscomment{Checked, Sep 1.}



\subsection{Pearson Divergence}
\label{sec:pd}

\noindent The last EMQ we discuss is the \emph{Pearson Divergence}
($\pd$ -- see \citep{du-Plessis:2012nr}), also called the
\emph{$\chi^{2}$ Divergence} \citep{Liese:2006qf}, and defined as
\begin{equation}
  \begin{aligned}
    \label{eq:pearson}
    \pd(p,\hat{p})
    = & \frac{1}{|\mathcal{C}|}\sum_{c\in \mathcal{C}}
    \frac{(p(c)-\hat{p}(c))^{2}}{\hat{p}(c)}
  \end{aligned}
\end{equation}
\noindent As an EMQ, $\pd$ has been first used by
\cite{Ceron:2016fk}. $\pd$ is undefined when, for a given class $c$,
$\hat{p}(c)$ is zero; the smoothed values of Equation
\ref{eq:smoothing} must thus be used within Equation \ref{eq:pearson}
in order to avoid this problem.

The arguments already used for $\aae$ in Appendix
\ref{sec:ThePropertiesOfAE} can be easily used to show that $\pd$
enjoys \textbf{IoI}, \textbf{NN}, and \textbf{IND}. That $\pd$ enjoys
\textbf{MON} is instead not self-evident; the proof that it indeed
does is reported in Appendix \ref{sec:provingmon}.

That $\pd$ enjoys neither \textbf{MAX}, nor \textbf{IMP}, nor
\textbf{REL}, nor \textbf{ABS}, is shown in Appendixes
\ref{sec:testingforMAX}, \ref{sec:testingforIMP},
\ref{sec:testingforREL}, \ref{sec:testingforABS}, respectively. The
fact that $\pd$ does not enjoy \textbf{MAX} can also be shown with
arguments used for showing the same for $\kld$; that is, when a
predicted probability $\hat{p}(c)$ is very small, $\pd$ becomes very
large.  Thanks to the fact that we use smoothed values, though,
$\hat{p}$ is lower-bounded by $\epsilon$, and $\pd$ has thus a finite
upper bound.  Like for other EMQs we have already discussed, the
perverse estimator for $\pd$ is the one that attributes 1 to the
probability of class $c^{*}=\arg\min_{c\in\mathcal{C}}p(c)$ and 0 to
the other classes, and its value is thus
\begin{equation}
  \label{eq:pdupperbound}
  z_{\pd}(p,\hat{p})=\frac{1}{|\mathcal{C}|}(\frac{1-(|\mathcal{C}|-1)\cdot \epsilon - p_{s}(c^{*})}{1-(|\mathcal{C}|-1)\cdot \epsilon}+\sum_{c\in\mathcal{C}/\{c^{*}\}}\frac{(p(c)-\epsilon)^{2}}{\epsilon})
\end{equation}
\noindent which shows that the range of $\pd$ depends on $p$, the
cardinality of $\mathcal{C}$, and the value of $\epsilon$. This
suffices to show that $\pd$ does not enjoy \textbf{MAX}.

\fabscomment{Checked, Sep 1.}



\section{Discussion}
\label{sec:discussion}

\noindent The properties that the EMQs of Section \ref{sec:measures}
enjoy and do not enjoy are conveniently summarized in Table
\ref{sec:propertiesofEMQs}. Table \ref{sec:adopters} lists instead the
papers where the various EMQs have been proposed and the papers where
they have subsequently been used for evaluation purposes. 

\begin{table}[tb]
  \caption{Properties of the EMQs discussed in this paper.}
  \begin{center}
    \begin{tabularx}{.90\textwidth}{ |r|| *{8}{Y|} }
      \hline
      & \textbf{IoI} &  \textbf{NN} & \textbf{MAX} & \textbf{MON} & \textbf{IMP} & \textbf{REL} & \textbf{ABS} & \textbf{IND} \\
      \hline\hline
      $\aae$   & Yes & Yes & \no & Yes & Yes & \no & Yes & Yes \\ \hline
      $\nae$   & Yes & Yes & Yes & Yes & Yes & \no & \no & Yes \\ \hline
      $\rae$   & Yes & Yes & \no & Yes & Yes & Yes & \no & Yes \\ \hline
      $\nrae$  & Yes & Yes & Yes & Yes & Yes & \no & \no & Yes \\ \hline
      $\se$    & Yes & Yes & \no & Yes & Yes & \no & Yes & Yes \\ \hline
      $\dr$    & Yes & Yes & \no & Yes & \no & Yes & \no & Yes \\ \hline
      $\kld$   & Yes & Yes & \no & Yes & \no & \no & \no & Yes \\ \hline
      $\nkld$  & Yes & Yes & Yes & Yes & \no & \no & \no & Yes \\ \hline
      $\pd$    & Yes & Yes & \no & Yes & \no & \no & \no & Yes \\ \hline
    \end{tabularx}
  \end{center}
  \label{sec:propertiesofEMQs}
\end{table}


\begin{table}[tb]
  \caption{Research works about quantification where the EMQs discussed in this paper have been first proposed ($\bigstar$) and later used (\checkmark).}
  \begin{center}
    \resizebox{\textwidth}{!}{
    \begin{tabularx}{\textwidth}{ |r|| *{9}{Y|} }
      \hline
      & \begin{sideways}$\aae$\end{sideways} & \begin{sideways}$\nae$\end{sideways} & \begin{sideways}$\rae$\end{sideways} & \begin{sideways}$\nrae$\end{sideways} & \begin{sideways}$\se$\end{sideways} & \begin{sideways}$\dr$\end{sideways} & \begin{sideways}$\kld$\end{sideways} & \begin{sideways}$\nkld$\phantom{00}\end{sideways} & \begin{sideways}$\pd$\end{sideways} \\ \hline\hline
      \citep{Saerens:2002uq} & $\bigstar$ &  &  &  &  &  &  &  & \\ \hline
      \citep{Forman:2005fk} & \checkmark &  &  &  &  &  & $\bigstar$ &  & \\ \hline
      \citep{Forman:2006uf} & \checkmark &  &  &  &  &  & \checkmark &  & \\ \hline
      \citep{Forman:2008kx} & \checkmark &  &  &  &  &  & \checkmark &  & \\ \hline
      \citep{Tang:2010uq} & \checkmark &  &  &  &  &  & \checkmark &  & \\ \hline
      \citep{Bella:2010kx} & \checkmark &  &  &  & $\bigstar$ &  &  &  & \\ \hline
      \citep{Gonzalez-Castro:2010fk} & \checkmark &  & $\bigstar$ &  &  &  &  &  & \\ \hline
      \citep{Zhang:2010kx} & \checkmark &  &  &  &  &  &  &  & \\ \hline
      \citep{Alaiz-Rodriguez:2011fk} & \checkmark &  & \checkmark &  &  &  &  &  & \\ \hline
      \citep{Milli:2013fk} &  &  &  &  &  &  & \checkmark &  & \\ \hline
      \citep{Barranquero:2013fk} & \checkmark &  &  &  &  &  &  &  & \\ \hline
      \citep{Gonzalez-Castro:2013fk} & \checkmark &  & \checkmark &  &  &  &  &  & \\ \hline
      \citep{Esuli:2014uq} & \checkmark & $\bigstar$ & \checkmark & $\bigstar$ &  &  & \checkmark & $\bigstar$ & \\ \hline
      \citep{du-Plessis:2014kl} &  &  &  &  & \checkmark &  &  &  & \\ \hline
      \citep{Esuli:2015gh} &  &  & \checkmark &  &  &  & \checkmark &  & \\ \hline
      \citep{Gao:2015ly} & \checkmark & \checkmark & \checkmark & \checkmark &  &  & \checkmark & \checkmark & \\ \hline
      \citep{Barranquero:2015fr} & \checkmark &  &  &  &  &  & \checkmark &  & \\ \hline
      \citep{Beijbom:2015yg} & \checkmark &  &  &  &  &  &  &  & \\ \hline
      \citep{Milli:2015mz} &  &  &  &  &  &  & \checkmark &  & \\ \hline
      \citep{Gao:2016uq} & \checkmark & \checkmark & \checkmark & \checkmark &  &  & \checkmark & \checkmark & \\ \hline
      \citep{Ceron:2016fk} & \checkmark &  &  &  &  &  &  &  & $\bigstar$ \\ \hline
      \citep{Gonzalez:2016xy} & \checkmark &  &  &  &  &  &  &  & \\ \hline
      \citep{Kar:2016rw} &  &  &  &  &  &  & \checkmark &  & \\ \hline
      \citep{Nakov:2016ty} &  &  &  &  &  &  & \checkmark &  & \\ \hline
      \citep{Plessis:2017sp} &  &  &  &  & \checkmark &  &  &  & \\ \hline
      \citep{Levin:2017dq} &  &  &  &  &  & $\bigstar$ &  &  & \\ \hline
      \citep{Perez-Gallego:2017wt} & \checkmark &  &  &  & \checkmark &  &  &  & \\ \hline
      \citep{Tasche:2017ij} &  &  & \checkmark &  &  &  &  &  & \\ \hline
      \citep{Nakov:2017ty} & \checkmark &  & \checkmark &  &  &  & \checkmark &  & \\ \hline
      \citep{Maletzke:2017fk} & \checkmark & \checkmark &  & \checkmark &  &  &  &  & \\ \hline
      \citep{Esuli:2018rm} & \checkmark &  & \checkmark &  &  &  & \checkmark &  & \\ \hline
      \citep{Card:2018pb} & \checkmark &  &  &  &  &  &  &  & \\ \hline
      \citep{Reis:2018fk} & \checkmark & &  & \checkmark &  &  &  &  & \\ \hline
      \citep{Fernandes-Vaz:2018rc} &  &  &  &  & \checkmark &  &  &  & \\ \hline
      \citep{Sanya:2018wt} &  &  &  &  &  &  & \checkmark &  & \\ \hline
      \citep{Perez-Gallego:2019vl} & \checkmark &  &  &  & \checkmark &  &  &  & \\
      \hline
    \end{tabularx}
    }
  \end{center}
  \label{sec:adopters}
\end{table}%


\subsection{Are all our Properties Equally Important?}
\label{sec:relativeimportance}

\noindent An examination of Table \ref{sec:propertiesofEMQs} allows us
to make a number of general considerations. The first one is that some
of our properties (namely: \textbf{IoI}, \textbf{NN}, \textbf{MON},
\textbf{IND}) are unproblematic, since all the EMQs proposed so far
satisfy them, while other properties (namely: \textbf{MAX},
\textbf{IMP}, \textbf{REL}, \textbf{ABS}) are failed by several EMQs,
including ones (e.g., $\aae$, $\kld$) that are almost standard in the
quantification literature.  The second, related observation is that,
if we agree on the fact that the eight properties we have discussed
are desirable, a number of EMQs that have been proposed in the
quantification literature emerge as severely inadequate, since they
fail several among these properties; this is true even if we discount
the fact that, as we have already observed, \textbf{REL} and
\textbf{ABS} are mutually exclusive. The case of $\kld$ (which fails
on counts of \textbf{MAX}, \textbf{IMP}, \textbf{REL}, \textbf{ABS})
is of special significance, since $\kld$ has almost become a standard
in the evaluation of single-label (and binary) quantification (from
Table \ref{sec:adopters} $\kld$ emerges as the 2nd most frequently
used EMQ, after $\aae$).

However, an even more compelling fact that emerges from Table
\ref{sec:propertiesofEMQs} is that no EMQ among those proposed so far
satisfies (even discounting the mutual exclusivity of \textbf{REL} and
\textbf{ABS}) all the proposed properties. This suggests that more
research is needed in order to identify, or synthesize, an EMQ more
satisfactory than all the existing ones.

At the same time, in the absence of a truly satisfactory EMD, we think
that it is important to analyse whether all of our properties are
equally important, or if some of them is less important than others
and can thus be ``sacrificed''. Judging from Table
\ref{sec:propertiesofEMQs}, the key stumbling block seems to be the
\textbf{MAX} property, since all the EMQs that satisfy \textbf{MAX}
(namely: $\nae$, $\nrae$, $\nkld$) satisfy neither \textbf{REL} nor
\textbf{ABS}. This is undesirable since, as argued at the end of
Section \ref{sec:the8properties}, some applications of quantification
do require \textbf{REL}, while some other applications do require
\textbf{ABS} (and we can think of no application that requires
neither). \fabscomment{Interestingly enough, all the EMQs that satisfy
\textbf{MAX} (namely: $\nae$, $\nrae$, $\nkld$) do not arise
``naturally'', but are the result of a normalization effort
specifically aimed at making an existing measure (namely: $\aae$,
$\rae$, $\kld$) satisfy \textbf{MAX}. $\aae$ satisfies \textbf{ABS},
$\rae$ satisfies \textbf{REL}, and after being normalized these
properties are lost. The reason why they are lost is the fact that the
maximum value that the unnormalized versions can have is not a
constant (otherwise these unnormalized versions would enjoy
\textbf{MAX} ...), and normalizing by this maximum value has the
effect ...} Among the EMQs that satisfy \textbf{ABS} (and not
\textbf{REL}), $\aae$ and $\se$ satisfy all other properties but
\textbf{MAX}, while among the ones that satisfy \textbf{REL} (and not
\textbf{ABS}), also $\rae$ satisfies all other properties but
\textbf{MAX}.

In other words, if we stick to available EMQs, if we want \textbf{ABS}
or \textbf{REL} we need to renounce to \textbf{MAX}, while if we want
\textbf{MAX} we need to renounce to both \textbf{ABS} and
\textbf{REL}. How relatively desirable are these three properties? We
recall from Section \ref{sec:the8properties} that
\begin{enumerate}

\item \label{item:rel} the argument in favour of \textbf{REL} is that
  it reflects the needs of applications in which an estimation error
  of a given absolute magnitude should be considered more serious if
  it affects a rarer class;

\item \label{item:abs} the argument in favour of \textbf{ABS} is that
  it reflects the needs of applications in which an estimation error
  of a given absolute magnitude has the same impact independently from
  the true prevalence of the affected class;

\item \label{item:max} the main (although not the only) argument in
  favour of \textbf{MAX} is that, if an EMD does not satisfy it, the
  $n$ samples on which we may want to compare our quantification
  algorithms will each have a different weight on the final result.

\end{enumerate}
\noindent The relative importance of these three arguments is probably
a matter of opinion. However, it is our impression that Arguments
\ref{item:rel} and \ref{item:abs} are more compelling than Argument
\ref{item:max}, since \ref{item:rel} and \ref{item:abs} are really
about how an evaluation measure reflects the needs of the application
for which one performs a given task (quantification, in our case); if
the corresponding properties are not satisfied, one may argue that the
quantification accuracy (or error) being measured is only loosely
related to what the user really wants. Argument \ref{item:max}, while
important, ``only'' implies that, if \textbf{MAX} is not satisfied,
some samples will weigh more than others on the final result; while
undesirable, this does not affect the experimental comparison among
different quantification systems, since each of them is affected by
this disparity in the same way.\footnote{A similar situation occurs
when evaluating multi-label classification via ``microaveraged
$F_{1}$'', a measure in which the classes with higher prevalence weigh
more on the final result.} So, if we accept the idea of
``sacrificing'' \textbf{MAX} in order to retain \textbf{REL} or
\textbf{ABS}, Table \ref{sec:propertiesofEMQs} indicates that our
measures of choice should be

\begin{itemize}

\item $\aae$ (or $\se$, which is structurally similar), for those
  applications in which an estimation error of a given absolute
  magnitude should be considered more serious when the true prevalence
  of the affected class is lower; and

\item $\rae$, for those applications in which an estimation error of a
  given absolute magnitude has the same impact independently from the
  true prevalence of the affected class.

\end{itemize}

\subsection{\blue{Properties that Escape Formalization}}
\label{sec:ineffable}

\noindent \blue{While all the above discussion on the properties of
EMQs has been unashamedly formal, we should also remember that
choosing an evaluation measure instead of another should also be
guided by practical considerations, i.e., by properties of the measure
that are not necessarily amenable to formalization.
One such property is understandability, i.e., how simple and intuitive
is the mathematical form of an evaluation measure. While such
simplicity might not be a primary concern for the researcher, or the
mathematician, it might be for the practitioner. For instance, a
company that wants to sell a text analytics product to a customer
might need to run experiments on the customer's own data and explain
the results to the customer; since customers might not be
mathematically savvy, the fact that the measure chosen is easily
understandable to people with a minimal mathematical background is
important. On this account, measures such as $\aae$ and $\rae$
certainly win over other measures such as $\kld$ and $\nkld$, which
the average customer would find hardly intelligible.\footnote{It is
this author's experience that even measures such as $F_{1}$ can be
considered by customers ``esoteric''.}}

\blue{Another property that is difficult to formalize is
\emph{robustness to outliers}. Many EMQs often take the form of an
average
$D(p,\hat{p})=\frac{1}{|\mathcal{C}|}\sum_{c\in
\mathcal{C}}f(p(c),\hat{p}(c))$ across the classes in the
codeframe. If $D(p,\hat{p})$ is not ``robust to outliers'', it means
that an extreme value $f(p(c'),\hat{p}(c'))$ that may occur for some
$c'\in \mathcal{C}$ dominates on all the other values
$f(p(c),\hat{p}(c))$ for $c\in\mathcal{C}/\{c'\}$, giving rise to a
high value of $D(p,\hat{p})$ that is essentially due to $c'$ only. As
the name implies, ``robustness to outliers'' is usually considered a
desirable property; however, in some contexts it might also viewed as
undesirable (e.g., we might want to avoid quantification methods that
generate blatant mistakes, so we might want a measure that penalizes
the presence of even one of them). Aside from the fact that its
desirability is questionable, it should also be mentioned that
``robustness to outliers'' comes in degrees. E.g., absolute error is
more robust to outliers than squared error, but squared error is more
robust to outliers than ``cubic error'', etc.; and all of them are
vastly more robust to outliers than $\kld$ and $\nkld$. Which among
these enforces the ``right'' level of robustness to outliers? This
shows that robustness to outliers, independently from its
desirability, cannot be framed as a binary property (i.e., one that a
measure either enjoys or not), and thus escapes the type of analysis
that we have carried out in this paper.

Another property which is difficult to formalize has to do with the
set of values which an EMQ ranges on when evaluating \emph{realistic}
quantification systems (i.e., systems that exhibit a quantification
accuracy equal or superior to, say, that of a trivial ``classify and
count'' approach using SVMs). For these systems, the actual values
that an EMQ takes should occupy a fairly small subinterval of its
entire range. The question is: how small? One particularly problematic
EMQ, from this respect, is $\kld$. While its range is $[0,z_{\kld}]$,
where $z_{\kld}$ is as in Equation \ref{eq:kldupperbound}, realistic
quantification systems generate \emph{very} small $\kld$ values, so
small that they are sometimes difficult to make sense of.  One result
is that two genuine quantifiers that are being compared experimentally
may easily obtain results several \emph{orders of magnitude}
away. Such differences in performance are difficult to
grasp.\footnote{\blue{As an example, assume a (very realistic)
scenario in which $|\sigma|=1000$, $\mathcal{C}=\{c_{1},c_{2}\}$,
$p(c_{1})=0.01$, and in which three different quantifiers $\hat{p}'$,
$\hat{p}''$, $\hat{p}'''$ are such that $\hat{p}'(c_{1})=0.0101$,
$\hat{p}''(c_{1})=0.0110$, $\hat{p}'''(c_{1})=0.0200$. In this
scenario $\kld$ ranges in $[0,7.46]$, $\kld(p,\hat{p}')=4.78$e-07,
$\kld(p,\hat{p}'')=4.53$e-05, $\kld(p,\hat{p}''')=3.02$e-03, i.e., the
difference between $\kld(p,\hat{p}')$ and $\kld(p,\hat{p}'')$ (and the
one between $\kld(p,\hat{p}'')$ and $\kld(p,\hat{p}''')$) is 2 orders
of magnitude, while the difference between $\kld(p,\hat{p}')$ and
$\kld(p,\hat{p}''')$ is no less than 4 orders of magnitude. The
increase in error (as computed by $\kld$) deriving from using
$\hat{p}'''$ instead of $\hat{p}'$ is +632599\%.}}  We should add to
this that, if one wants to average $\kld$ results across a set of
samples (on this see also Section \ref{sec:multiplesamples}), the
average is completely dominated by the value with the highest order of
magnitude, and the others have little or no impact. Unfortunately,
switching from $\kld$ to $\nkld$ does not help much in this respect
since, for realistic quantification systems,
$\nkld(p,\hat{p})\approx\frac{1}{2}\kld(p,\hat{p})$. The reason is
that $\nkld$ is obtained by applying a sigmoidal function (namely, the
logistic function) to $\kld$, and the tangent to this sigmoid for
$x=0$ is $y=\frac{1}{2}x$; since the values of $\kld$ for realistic
quantifiers are (as we have observed above) very close to 0, for these
values the $\nkld(p,\hat{p})$ curve is well approximated by
$y=\frac{1}{2}\kld(p,\hat{p})$. As an EMQ, $\nkld$ thus \emph{de
facto} inherits most of the problems of $\kld.$
 
All of the above points to the fact that choosing a good EMQ (and the
same may well be true for tasks other than quantification) should also
be based, aside from the formal properties that the EMQ enjoys, on
criteria that either resist or completely escape formalization, such
as understandability and ease of use.}

\fabscomment{Checked, Sep 3.}


\subsection{Evaluating Quantification across Multiple Samples}
\label{sec:multiplesamples}

\noindent On a different note, we also need to stress a key difference
between measures of classification accuracy and measures of
quantification accuracy (or error). The objects of classification are
individual unlabelled items, and all measures of classification
accuracy (e.g., $F_{1}$) are defined with respect to a test \emph{set}
of such objects. The objects of quantification, instead, are samples,
and all the measures of quantification accuracy we have discussed in
this paper are defined on a \emph{single} such sample (i.e., they
measure how well the true distribution of the classes \emph{across
this individual sample} is approximated by the predicted distribution
of the classes across the same sample). Since every evaluation is
worthless if carried out on a single object, it is clear that
quantification systems need to be evaluated on \emph{sets} of
samples. This means that every measure that we have discussed needs
first to be evaluated on each sample, and then its global score across
the test set (i.e. the set of samples on which testing is carried out)
needs to be computed. This global score may be computed via any
measure of central tendency, e.g., via an average, or a median, or
other (for instance, if $\nae$ is used, we might in turn use
\emph{Average $\nae$} or \emph{Median $\nae$}, where averages and
medians are computed across a set of samples). We do not take any
specific stand for or against computing global scores via any specific
measure of central tendency, since each of them may serve different
but legitimate purposes. \fabscomment{Checked, Sep 3.}


\section{Conclusions}
\label{sec:conclusion}

\noindent We have presented a study that ``evaluates evaluation'', in
the tradition of the so-called ``axiomatic'' approach to the study of
evaluation measures for information retrieval and related tasks. Our
effort has targeted quantification, an important task at the
crossroads of information retrieval, data mining, and machine
learning, and has consisted of analysing previously proposed
evaluation measures for quantification using the toolbox of the
above-mentioned ``axiomatic'' approach. The work closest in spirit to
the present one is our past work on the analysis of evaluation
measures for classification \citep{Sebastiani:2015zl}. However,
quantification poses more difficult problems than classification,
since evaluation measures for quantification are inherently nonlinear
(i.e., quantification error cannot be expressed as a linear function
of the labelling error made on individual items). This is unlike
classification, for which linear measures (e.g., standard accuracy, or
$K$ -- see \cite{Sebastiani:2015zl}) are possible.

We have proposed eight properties that, as we have argued, are
desirable for measures that attempt to evaluate quantification (two
such properties are actually mutually exclusive, and are desirable
each in a different class of applications of quantification). Our
analysis has revealed that, unfortunately, no existing evaluation
measure for quantification satisfies all of these properties. While
this points to the fact that more research is needed to identify, or
synthesize, a truly adequate such measure, this also means that, for
the moment being, we have to evaluate the relative desirability of the
properties that the existing measures do not satisfy. We have argued
that some such properties are more important than others, and that as
a result two measures (``Absolute Error'' and ``Relative Absolute
Error'') stand out as the most satisfactory ones (interestingly
enough, they are also the most time-honoured ones, and the
mathematically simplest ones). As we have argued, the former is more
adequate for application contexts in which an estimation error of a
given absolute magnitude should be considered more serious if it
affects a rare class, while the latter is more adequate for those
applications in which an estimation error of a given absolute
magnitude has the same impact independently from the true prevalence
of the affected class. \fabscomment{Checked, Sep 3.}


\bibliographystyle{ACM-Reference-Format-Journals}
\bibliography{Fabrizio}


\appendix


\section{Properties of $\aae$}
\label{sec:ThePropertiesOfAE}

\noindent We here prove that $\aae$ enjoys \textbf{IoI}, \textbf{NN},
\textbf{IND}, \textbf{MON}, \textbf{IMP}, \textbf{ABS}. While some of
these proofs are trivial, these are reported in detail in order to
show how the same arguments can be used to prove the same for many of
the other EMQs discussed in Section \ref{sec:measures}.

$\aae$ enjoys \textbf{IoI}. In fact,
$\aae(p,\hat{p})=\frac{1}{|\mathcal{C}|}\sum_{c\in
\mathcal{C}}|\hat{p}(c)-p(c)|=0$ implies that
$\sum_{c\in \mathcal{C}}|\hat{p}(c)-p(c)|=0$; given that
$\sum_{c\in \mathcal{C}}|\hat{p}(c)-p(c)|$ is a sum of nonnegative
factors, this implies that $|\hat{p}(c)-p(c)|=0$ for all
$c\in\mathcal{C}$, i.e., $\hat{p}(c)=p(c)$ for all
$c\in\mathcal{C}$. Conversely, if $\hat{p}=p$, then
$\frac{1}{|\mathcal{C}|}\sum_{c\in
\mathcal{C}}|\hat{p}(c)-p(c)|=0$. \hspace{3em}\mbox{}\qed

$\aae$ enjoys \textbf{NN}. Quite obviously,
$\frac{1}{|\mathcal{C}|}\geq 0$ and
$\sum_{c\in \mathcal{C}}|\hat{p}(c)-p(c)|\geq 0$, which implies that
$\frac{1}{|\mathcal{C}|}\sum_{c\in \mathcal{C}}|\hat{p}(c)-p(c)|\geq
0$. \hspace{3em}\mbox{}\qed

$\aae$ enjoys \textbf{IND}. Given codeframe
$\mathcal{C}=\{c_{1}, ..., c_{k}, c_{k+1}, ..., c_{n}\}$, for any true
distribution $p$ on $\mathcal{C}$ and predicted distributions
$\hat{p}'$ and $\hat{p}''$ on $\mathcal{C}$ such that
$\hat{p}'(c)= \hat{p}''(c)$ for all $c\in\{c_{k+1}, ..., c_{n}\}$, the
inequality
$$\aae(p,\hat{p}')\leq \aae(p,\hat{p}'')$$ resolves to 
\begin{equation}\nonumber
  \begin{aligned}
    \frac{1}{|\mathcal{C}|}\sum_{c\in \mathcal{C}}|\hat{p}'(c)-p(c)|\leq & \ \frac{1}{|\mathcal{C}|}\sum_{c\in \mathcal{C}}|\hat{p}''(c)-p(c)| \\
    \sum_{c\in \mathcal{C}}|\hat{p}'(c)-p(c)|\leq & \ \sum_{c\in \mathcal{C}}|\hat{p}''(c)-p(c)| \\
    \sum_{c\in \mathcal{C}_{1}}|\hat{p}'(c)-p(c)|+\sum_{c\in \mathcal{C}_{2}}|\hat{p}'(c)-p(c)| \leq & \ \sum_{c\in \mathcal{C}_{1}}|\hat{p}''(c)-p(c)|+\sum_{c\in \mathcal{C}_{2}}|\hat{p}''(c)-p(c)| \\
    \sum_{c\in \mathcal{C}_{1}}|\hat{p}'(c)-p(c)| \leq  & \ \sum_{c\in \mathcal{C}_{1}}|\hat{p}''(c)-p(c)| \\
    \frac{1}{|\mathcal{C}_{1}|}\sum_{c\in \mathcal{C}_{1}}|\hat{p}'(c)-p(c)| \leq  & \ \frac{1}{|\mathcal{C}_{1}|}\sum_{c\in \mathcal{C}_{1}}|\hat{p}''(c)-p(c)| \\
    \aae(p_{\mathcal{C}_{1}},\hat{p}_{\mathcal{C}_{1}}')\leq & \
    \aae(p_{\mathcal{C}_{1}},\hat{p}_{\mathcal{C}_{1}}'')
    \hspace{3em}\mbox{}\qed
  \end{aligned}
\end{equation}
$\aae$ enjoys \textbf{MON}. This can be proven by showing that $\aae$
enjoys \textbf{B-MON}, since we have proven that it enjoys
\textbf{IND}. Given codeframe $\mathcal{C}=\{c_{1},c_{2}\}$ and true
distribution $p$, if predicted distributions $\hat{p}',\hat{p}''$ are
such that $\hat{p}''(c_{1})< \hat{p}'(c_{1})\leq p(c_{1})$, then it
holds that
\begin{equation}\nonumber
  \begin{aligned}
    \aae(p,\hat{p}')= & \ \frac{1}{2}(|\hat{p}'(c_{1})-p(c_{1})|+|\hat{p}'(c_{2})-p(c_{2})|) \\
    < & \ \frac{1}{2}(|\hat{p}''(c_{1})-p(c_{1})|+|\hat{p}''(c_{2})-p(c_{2})|) \\
    = & \ \aae(p,\hat{p}'') \hspace{3em}\mbox{}\qed
  \end{aligned}
\end{equation}
\noindent $\aae$ enjoys \textbf{IMP}. This can be shown by showing
that $\aae$ enjoys \textbf{B-IMP}, since we have proven that it enjoys
\textbf{IND}. Given codeframe $\mathcal{C}=\{c_{1},c_{2}\}$, true
distribution $p$, predicted distributions $\hat{p}'$ and $\hat{p}''$,
and constant $a\geq 0$ such that $\hat{p}'(c_{1})=p(c_{1})+a$ and
$\hat{p}''(c_{1})=p(c_{1})-a$, it holds that
\begin{equation}\nonumber
  \begin{aligned}
    \aae(p,\hat{p}')= & \ \frac{1}{2}(|\hat{p}'(c_{1})-p(c_{1})|+|\hat{p}'(c_{2})-p(c_{2})|) \\
    = & \
    \frac{1}{2}(|(p(c_{1})+a)-p(c_{1})|+|(p(c_{2})-a)-p(c_{2})|) \\
    = & \
    \frac{1}{2}(|a|+|-a|) \\
    = & \
    \frac{1}{2}(|-a|+|a|) \\
    = & \
    \frac{1}{2}(|(p(c_{1})-a)-p(c_{1})|+|(p(c_{2})+a)-p(c_{2})|) \\
    = & \ \aae(p,\hat{p}'') \hspace{3em}\mbox{}\qed
  \end{aligned}
\end{equation}
\noindent $\aae$ enjoys \textbf{ABS}. This can be shown by showing
that $\aae$ enjoys \textbf{B-ABS}, since we have proven that it enjoys
\textbf{IND}. Given codeframe $\mathcal{C}=\{c_{1},c_{2}\}$, constant
$a>0$, true distributions $p'$ and $p''$ such that
$p'(c_{1})<p''(c_{1})$ and $p''(c_{1})<p''(c_{2})$, if a predicted
distribution $\hat{p}'$ that estimates $p'$ is such that
$\hat{p}'(c_{1})=p'(c_{1})\pm a$ and a predicted distribution
$\hat{p}''$ that estimates $p''$ is such that
$\hat{p}''(c_{1})=p''(c_{1})\pm a$, then it holds that
\begin{equation}\nonumber
  \begin{aligned}
    \aae(p',\hat{p}')= & \ \frac{1}{2}(|\hat{p}'(c_{1})-p'(c_{1})|+|\hat{p}'(c_{2})-p'(c_{2})|) \\
    = & \ \frac{1}{2}(|(p'(c_{1})\pm a)-p'(c_{1})|+|(p'(c_{2})\mp a)-p'(c_{2})|) \\
    = & \ \frac{1}{2}(2a)\\
    = & \ \frac{1}{2}((p''(c_{1})\pm a)-p''(c_{1})|+|(p''(c_{2})\mp a)-p''(c_{2})|) \\
    = & \ \frac{1}{2}(|\hat{p}''(c_{1})-p''(c_{1})|+|\hat{p}''(c_{2})-p''(c_{2})|) \\
    = & \ \aae(p'',\hat{p}'') \hspace{3em}\mbox{}\qed
  \end{aligned}
\end{equation}


\section{Testing for \textbf{MAX}, \textbf{IMP}, \textbf{ABS},
\textbf{REL}}
\label{sec:testing}

\noindent In this section we present simple tests aimed at
establishing that a certain EMQ $D$ does \emph{not} enjoy a certain
property
$\pi\in\{\textbf{MAX}, \textbf{IMP}, \textbf{ABS},
\textbf{REL}\}$. The basic pattern of these tests is to show that
$\pi$ does not hold for $D$ by providing a counterexample. More in
particular, given a concrete scenario $s$ characterized by (1) a
codeframe $\mathcal{C}$, (2) one or more true distributions $p_{1}$,
$p_{2}$, \ldots, and (3) one or more predicted distributions
$\hat{p}_{1}$, $\hat{p}_{2}$, \ldots, the test attempts to check
whether the scenario satisfies the particular constraint that is
required for property $\pi$ to hold. Since for $D$ to enjoy property
$\pi$ the constraint is required to hold for all scenarios, if $\pi$
does not hold in $s$ we can conclude than $D$ does not enjoy
$\pi$. Instead, if $\pi$ does hold in $s$ we can conclude nothing, and
thus need to study the issue further.


\subsection{A Counterexample for \textbf{MAX}}
\label{sec:testingforMAX}

\noindent In the test for \textbf{MAX} we consider the scenario
described in the following table
\begin{center}
  \resizebox{\textwidth}{!} {
  \begin{tabular}{|l|c|c|c|c||c|c|c|c|c|c|c|c|c|c|c|c|}
    \hline
    & \side{$p(c_{1})$} & \side{$p(c_{2})$} & \side{$\hat{p}(c_{1})$} & \side{$\hat{p}(c_{2})$} & \side{$\aae$} & \side{$\nae$} & \side{$\rae$} & \side{$\nrae$ \hspace{.5ex}\mbox{}}  & \side{$\se$} & \side{$\dr$} & \side{$\kld$} & \side{$\nkld$} & \side{$\pd$} \\
    \hline
    $p'$  & 0.01 & 0.99 & 1.00 & 0.00 & \gr 0.9900 & 1.0000 & \gr 49.9975 & 1.0000 & \gr 0.9801 & \gr 0.9950 & \gr 14.3076 & 0.9999 & \gr 980100.0004 \\
    $p''$ & 0.49 & 0.51 & 1.00 & 0.00 & \gr 0.5100 & 1.0000 & \gr \phantom{4}1.0204 & 1.0000 & \gr 0.2601 & \gr 0.7550 & \gr \phantom{1}6.7065 & 0.9975 & \gr 260100.0001 \\
    \hline
  \end{tabular}
  }
\end{center}
\noindent and characterized by two different true distributions (1st
and 2nd row) across the same codeframe
$\mathcal{C}=\{c_{1},c_{2}\}$.\footnote{We assume
$|D|=1,000,000$. This assumption has no relevance on the qualitative
conclusions we draw here, and only affects the magnitude of the values
in the table (since the value of $|D|$ affects the value of
$\epsilon$, and thus of $\rae$, $\nrae$, $\dr$, $\kld$, $\nkld$, $\pd$
-- see Section \ref{sec:rae}) and following.} The test consists in
checking whether their respective perverse estimators obtain from $D$
the same score: if the values of measure $D$ in the two rows are not
the same (greyed-out cells), this implies that $D$ does not satisfy
\textbf{MAX} (if they are the same, this does \emph{not} necessarily
mean that $D$ satisfies \textbf{MAX}). Concerning the values obtained
by $\nkld$, see the discussion in Footnote \ref{foot:nkld}.

The table shows that none of $\aae$, $\rae$, $\se$, $\dr$, $\kld$,
$\pd$ satisfies \textbf{MAX}. \fabscomment{Checked, Aug 31.}


\subsection{A Counterexample for \textbf{IMP}}
\label{sec:testingforIMP}

\noindent In the test for \textbf{IMP} we consider the scenario
described in the following table
\begin{center}
  \resizebox{\textwidth}{!} {
  \begin{tabular}{|l||c|c||c|c||c|c|c|c|c|c|c|c|c|c|c|c|}
    \hline
    & \side{$p(c_{1})$} & \side{$p(c_{2})$} & \side{$\hat{p}(c_{1})$} & \side{$\hat{p}(c_{2})$} & \side{$\aae$} & \side{$\nae$} & \side{$\rae$} & \side{$\nrae$ \hspace{.5ex}\mbox{}}  & \side{$\se$} & \side{$\dr$} & \side{$\kld$} & \side{$\nkld$} & \side{$\pd$} \\
    \hline\hline
    $p'$  & 0.20 & 0.80 & 0.25 & 0.75 & 0.0500 & 0.0625 & 0.1562 & 0.0625 & 0.0025 & \gr 0.1312 & \gr 0.0070 & \gr 0.0035 & \gr 0.0117 \\
    $p''$ & 0.20 & 0.80 & 0.15 & 0.85 & 0.0500 & 0.0625 & 0.1562 & 0.0625 & 0.0025 & \gr 0.1544 & \gr 0.0090 & \gr 0.0045 & \gr 0.0181 \\
    \hline
  \end{tabular}
  }
\end{center}
\noindent and characterized by a codeframe
$\mathcal{C}=\{c_{1},c_{2}\}$, a true distribution $p$ (Columns 2 and
3), and two predicted distributions $\hat{p}'$ and $\hat{p}''$
(Columns 4 and 5, Rows 2 and 3) which are such that (i) $\hat{p}'$
overestimates and $\hat{p}''$ underestimates the prevalence of a class
$c_{1}$ by a certain amount $a>0$ (here: 0.05), and, symmetrically,
(ii) $\hat{p}'$ overestimates and $\hat{p}''$ underestimates the
prevalence of another class $c_{2}$ by the same amount $a$. If the
values of $D(p,\hat{p}')$ and $D(p,\hat{p}'')$ are not the same (which
in the table is indicated by greyed-out cells), this implies that $D$
does not satisfy \textbf{IMP} (if they are the same, this does
\emph{not} necessarily mean that $D$ satisfies \textbf{IMP}).

The table shows that none of $\dr$, $\kld$, $\nkld$, $\pd$ satisfies
\textbf{IMP}. \fabscomment{Checked, Aug 31.}


\subsection{A Counterexample for \textbf{REL}}
\label{sec:testingforREL}

\noindent In the test for \textbf{REL} we consider the scenario
described in the following table
\begin{center}
  \resizebox{\textwidth}{!} {
  \begin{tabular}{|l||c|c||c|c||c|c|c|c|c|c|c|c|c|c|c|c|}
    \hline
    & \side{$p(c_{1})$} & \side{$p(c_{2})$} & \side{$\hat{p}(c_{1})$} & \side{$\hat{p}(c_{2})$} & \side{$\aae$} & \side{$\nae$} & \side{$\rae$} & \side{$\nrae$ \hspace{.5ex}\mbox{}}  & \side{$\se$} & \side{$\dr$} & \side{$\kld$} & \side{$\nkld$} & \side{$\pd$} \\
    \hline\hline
    $p'$  & 0.20 & 0.80 & 0.70 & 0.30 & \gr 0.5000 & \gr 0.6250 & 1.5625 & \gr 0.6250 & \gr 0.2500 & 0.6696 & \gr 0.5341 & \gr 0.2609 & \gr 0.7738 \\
    $p''$ & 0.25 & 0.75 & 0.75 & 0.25 & \gr 0.5000 & \gr 0.6667 & 1.3333 & \gr 0.6667 & \gr 0.2500 & 0.6667 & \gr 0.5493 & \gr 0.2679 & \gr 0.8333 \\
    \hline
  \end{tabular}
  }
\end{center}
\noindent with a codeframe $\mathcal{C}=\{c_{1},c_{2}\}$, two true
distributions $p'$ and $p''$ (Rows 2 and 3, Columns 2 to 4), and two
corresponding predicted distributions $\hat{p}'$ and $\hat{p}''$ (Rows
2 and 3, Columns 5 to 7), such that in both cases the predicted
distribution overestimates the prevalence of $c_{1}$ by the same
amount $a>0$ (here: 0.50), with $p'(c_{1})<p''(c_{1})$. Here, if it is
not the case that $D(p,\hat{p}')>D(p,\hat{p}'')$ (which in the table
is indicated by greyed-out cells), then $D$ does not satisfy
\textbf{REL} (if $D(p,\hat{p}')\not =D(p,\hat{p}'')$, this does
\emph{not} necessarily mean that $D$ satisfies \textbf{REL}).

The table shows that none of $\aae$, $\nae$, $\nrae$, $\se$, $\kld$,
$\nkld$, $\pd$ satisfies \textbf{REL}. \fabscomment{Checked, Aug 31.}


\subsection{A Counterexample for \textbf{ABS}}
\label{sec:testingforABS}

\noindent In the test for \textbf{ABS} we consider the same scenario
as described in Appendix \ref{sec:testingforREL}, i.e.,
\begin{center}
  \resizebox{\textwidth}{!} {
  \begin{tabular}{|l||c|c||c|c||c|c|c|c|c|c|c|c|c|c|c|c|}
    \hline
    & \side{$p(c_{1})$} & \side{$p(c_{2})$} & \side{$\hat{p}(c_{1})$} & \side{$\hat{p}(c_{2})$} & \side{$\aae$} & \side{$\nae$} & \side{$\rae$} & \side{$\nrae$ \hspace{.5ex}\mbox{}}  & \side{$\se$} & \side{$\dr$} & \side{$\kld$} & \side{$\nkld$} & \side{$\pd$} \\
    \hline\hline
    $p'$  & 0.20 & 0.80 & 0.70 & 0.30 & 0.5000 & \gr 0.6250 & \gr 1.5625 & \gr 0.6250 & 0.2500 & \gr 0.6696 & \gr 0.5341 & \gr 0.2609 & \gr 0.7738 \\
    $p''$ & 0.25 & 0.75 & 0.75 & 0.25 & 0.5000 & \gr 0.6667 & \gr 1.3333 & \gr 0.6667 & 0.2500 & \gr 0.6667 & \gr 0.5493 & \gr 0.2679 & \gr 0.8333 \\
    \hline
  \end{tabular}
  }
\end{center}
\noindent with a codeframe $\mathcal{C}=\{c_{1},c_{2}\}$, two true
distributions $p'$ and $p''$ (Rows 2 and 3, Columns 2 to 4), and two
corresponding predicted distributions $\hat{p}'$ and $\hat{p}''$ (Rows
2 and 3, Columns 5 to 7), such that in both cases the predicted
distribution overestimates the prevalence of $c_{1}$ by the same
amount $a>0$ (here: 0.50), with $p'(c_{1})<p''(c_{1})$. Here, if the
values of $D(p,\hat{p}')$ and $D(p,\hat{p}'')$ are not equal (which in
the table is indicated by greyed-out cells), this implies that $D$
does not satisfy \textbf{ABS} (if $D(p,\hat{p}')=D(p,\hat{p}'')$, this
does \emph{not} necessarily mean that $D$ satisfies \textbf{ABS}).

The table shows that none of $\nae$, $\rae$, $\nrae$, $\dr$, $\kld$,
$\nkld$, $\pd$ satisfies \textbf{ABS}. \fabscomment{Checked, Aug 31.}


\section{Proving that \textbf{MON} Holds}
\label{sec:provingmon}

\noindent In this section we prove that \textbf{MON} holds for $\kld$
and $\pd$. For this it will be sufficient to prove that $\kld$ and
$\pd$ enjoy \textbf{B-MON}, since it is immediate to verify that
$\kld$ and $\pd$ enjoy \textbf{IND}.

For ease of exposition, let us define the shorthands
$a\equiv p(c_{1})$ and $x\equiv \hat{p}(c_{1})$.\footnote{For the EMQs
that require smoothed probabilities to be used, these definitions
obviously need to be replaced by $a\equiv p_{s}(c_{1})$ and
$x\equiv \hat{p}_{s}(c_{1})$.} In order to show that $D$ satisfies
\textbf{B-MON} it is sufficient to show that
\begin{enumerate}

\item if $(a-x)>0$, then $\dfrac{\partial}{\partial (a-x)} D>0$ for
  $a,x,(a-x)\in(0,1)$

\item if $(x-a)>0$, then $\dfrac{\partial}{\partial (x-a)} D>0$ for
  $a,x,(x-a)\in(0,1)$

\end{enumerate}
\noindent because an increase in $(a-x)=(p(c_{1})-\hat{p}(c_{1}))$
implies an equivalent increase in $(p(c_{2})-\hat{p}(c_{2}))$ (same
for $(x-a)$).


\begin{theorem}
  \label{th:kldenjoysmon}$\kld$ satisfies \textbf{B-MON}.\end{theorem}

\noindent \textsc{Proof.}  We first treat the case $(a-x)>0$; let us
define $y\equiv(a-x)$. In this case
\begin{equation}
  \begin{aligned}
    \nonumber
    \dfrac{\partial}{\partial y}  \kld = & \ \dfrac{\partial}{\partial y}  (a \log\dfrac{a}{x}+(1-a)\log\dfrac{1-a}{1-x})\\
    = & \ \dfrac{\partial}{\partial y}  (a \log\dfrac{a}{a-y}+(1-a)\log\dfrac{1-a}{1-a+y}) \\
    = & \ \dfrac{-y}{(a-y-1)(a-y)} \\
    = & \ \dfrac{x-a}{(x-1)x} \\
  \end{aligned}
\end{equation}
\noindent Since we are in the case in which $(x-a)<0$, and since
$(x-1)<0$ and $x>0$, then $\dfrac{x-a}{(x-1)x}>0$ for all
$a,x,(a-x)\in(0,1)$.

Let us now treat the case $(x-a)>0$, and let us define
$y\equiv(x-a)$. In this case
\begin{equation}
  \begin{aligned}
    \nonumber
    \dfrac{\partial}{\partial y}  \kld = & \ \dfrac{\partial}{\partial y}  (a \log\dfrac{a}{x}+(1-a)\log\dfrac{1-a}{1-x})\\
    = & \ \dfrac{\partial}{\partial y} (a \log\dfrac{a}{y-a}+(1-a)\log\dfrac{1-a}{1-y+a}) \\
    = & \ \dfrac{-y}{(a+y-1)(a+y)} \\
    = & \ \dfrac{a-x}{(x-1)x} \\
  \end{aligned}
\end{equation}
\noindent Since in this case it holds that $(a-x)<0$, and since
$(x-1)<0$ and $x>0$, then $\dfrac{x-a}{(x-1)x}>0$ for all
$a,x,(x-a)\in(0,1)$. This concludes our proof. \hfill \mbox{} \qed


\begin{theorem}$\pd$ satisfies \textbf{B-MON}.\end{theorem}

\noindent \textsc{Proof.}  We first treat the case $(a-x)>0$; let us
define $y\equiv(a-x)$. In this case
\begin{equation}
  \begin{aligned}
    \nonumber
    \dfrac{\partial}{\partial y} \pd = & \ \dfrac{\partial}{\partial y}  (\dfrac{(a-x)^{2}}{x}+\dfrac{((1-a)-(1-x))^{2}}{1-x})\\
    = & \ \dfrac{\partial}{\partial y}  (\dfrac{y^{2}}{a-y}+\dfrac{((1-a)-(1-a+y))^{2}}{1-a+y})\\
    = & \ \dfrac{y(-2a^{2}+2a(y+1)-y)}{(a-y)^{2}(-a+y+1)^{2}} \\
    = & \ \dfrac{(a-x)(a-2ax+x)}{x^{2}(1-x)^{2}}
  \end{aligned}
\end{equation}
\noindent Since in this case it holds that $a>x$, it is true that that
$(a-2ax+x)>(x-2ax+x)=2x(1-a)>0$, since by hypothesis it holds that
$x,a\in (0,1)$. Therefore,
$\dfrac{\partial}{\partial y}
\pd=\dfrac{(a-x)(a-2ax+x)}{x^{2}(1-x)^{2}}>0$, since the two factors
at the numerator and the two factors at the denominator are all
strictly $>0$.

Let us now treat the case $(x-a)>0$, and let us define
$y\equiv(x-a)$. In this case
\begin{equation}
  \begin{aligned}
    \nonumber
    \dfrac{\partial}{\partial y}  \pd= & \ \dfrac{\partial}{\partial y}  (\dfrac{(a-x)^{2}}{x}+\dfrac{((1-a)-(1-x))^{2}}{1-x}) \\
    = & \ \dfrac{\partial (\dfrac{y^{2}}{y+a}+\dfrac{((1-a)-(1-a-y))^{2}}{1-a-y}}{\partial y}) \\
    = & \ \dfrac{y(-2a^{2}-2a(y-1)+y)}{(a-y+1)^{2}(a+y)^{2}} \\
    = & \ \dfrac{(x-a)(-2ax+x+a)}{(2a-x+1)^{2}x^{2}}
  \end{aligned}
\end{equation}
\noindent Since in this case it holds that $x>a$, it is true that that
$(-2ax+x+a)>(-2ax+2a)=2a(1-x)>0$, since by hypothesis it holds that
$x,a\in (0,1)$. Therefore,
$\dfrac{\partial \pd}{\partial
y}=\dfrac{(x-a)(-2ax+x+a)}{(2a-x+1)^{2}x^{2}}>0$, since the two
factors at the numerator and the two factors at the denominator are
all strictly $>0$. This concludes our proof. \hfill \mbox{} \qed

\end{document}